\documentclass{article}

\PassOptionsToPackage{numbers, compress}{natbib}

\usepackage[final]{neurips_2025}

\usepackage[utf8]{inputenc} 
\usepackage[T1]{fontenc}    
\usepackage{hyperref}       
\usepackage{url}            
\usepackage{booktabs}       
\usepackage{amsfonts}       
\usepackage{nicefrac}       
\usepackage{microtype}      
\usepackage{xcolor}         

\usepackage{graphicx}
\usepackage{multirow, multicol}
\usepackage{mathtools}
\usepackage{amsmath,amsfonts,amssymb,amsthm,bbm}
\usepackage{mathabx}
\usepackage{algorithm}
\usepackage{algorithmic}
\usepackage{colortbl}
\usepackage{amssymb}
\definecolor{mygray}{gray}{0.9}
\usepackage{wrapfig} 
\usepackage{threeparttable} 
\usepackage{caption}
\usepackage{makecell}
\usepackage{pifont}
\usepackage{enumitem}
\usepackage{subcaption}
\usepackage{longtable}
\usepackage[table]{xcolor}
\usepackage{soul}

\newcommand{\dataset}{{EgoBlind}}

\newcommand{\cmark}{\ding{51}}
\newcommand{\xmark}{\ding{55}}

\newcommand{\eg}{\textit{e.g.}}
\newcommand{\ie}{\textit{i.e.}}
\newcommand{\vs}{\textit{vs. }}

\title{EgoBlind: Towards Egocentric Visual Assistance for the Blind}

\author{
Junbin Xiao\textsuperscript{1}\thanks{Equal Contribution.},
\quad Nanxin Huang\textsuperscript{2*},
\quad Hao Qiu\textsuperscript{2},  
\quad Zhulin Tao\textsuperscript{2}\thanks{Corresponding Authors.}, \\
\quad \textbf{Xun Yang}\textsuperscript{3}, 
\quad \textbf{Richang Hong}\textsuperscript{4}, 
\quad \textbf{Meng Wang}\textsuperscript{4},  
\quad \textbf{Angela Yao}\textsuperscript{1$\dagger$} \\
\textsuperscript{1} National University of Singapore, \textsuperscript{2} Communication University of China  \\ 
\textsuperscript{3} University of Science and Technology of China, \textsuperscript{4} Hefei University of Technoloy  \\
}

\begin{document}
\maketitle

\begin{abstract}
We present EgoBlind, the first egocentric VideoQA dataset collected from blind individuals to evaluate the assistive capabilities of contemporary multimodal large language models (MLLMs). EgoBlind comprises 1,392 first-person videos from the daily lives of blind and visually impaired individuals. It also features 5,311 questions directly posed or verified by the blind to reflect their in-situation needs for visual assistance. Each question has an average of 3 manually annotated reference answers to reduce subjectiveness.
Using EgoBlind, we comprehensively evaluate 16 advanced MLLMs and find that all models struggle.  The best performers achieve an accuracy near 60\%, which is far behind human performance of 87.4\%. To guide future advancements, we identify and summarize major limitations of existing MLLMs in egocentric visual assistance for the blind and explore heuristic solutions for improvement. With these efforts, we hope that EgoBlind will serve as a foundation for developing effective AI assistants to enhance the independence of the blind and visually impaired. 
Data and code are available at \url{https://github.com/doc-doc/EgoBlind}.
\end{abstract}
\section{Introduction}
\label{sec:intro}
The rapid advancement of multimodal large language models (MLLMs) \cite{alayrac2022flamingo,li2023blip,liu2024visual,wang2024qwen2,hurst2024gpt,team2023gemini} has significantly improved the performance of visual question answering (VQA). However, existing VQA datasets primarily focus on the third-person perspective \cite{goyal2017making,xiao2021next} or general-purpose image and video understanding \cite{jia2022egotaskqa,cheng2024videgothink,mangalam2023egoschema,yang2021deconfounded,yang2022video,shang2019annotating}.  
Applications such as visual assistance for the visually-impaired \cite{gurari2018vizwiz}, who constitute an overwhelming 
2.2 billion people around the world \cite{whovision2019} have been received less attention. 
Research in assisting the blind from a first-person perspective is especially scarce \cite{song2024video}. 

In light of this, we construct \dataset~-- the first egocentric VideoQA dataset designed 
to benchmark and advance MLLMs towards egocentric visual assistance for the blind. \dataset~comprises 1,392 egocentric videos that capture the visual experiences of blind individuals and 5,311 questions that are directly posed or automatically generated and verified by blind users to reflect their assistive needs in exploring their surroundings. To categorize these needs for better analysis, the questions are grouped into six key types:
``Information Reading'', ``Safety Warning'', ``Navigation'',``Social Communication'', ``Tool Use'', and ``Other Resources''. We set the QA task as online (timestamp restricted) and open-form answer generation to better align with its live assistance nature. Multiple reference answers with well-aligned evaluation prompts are also provided for effective assessment.
\begin{figure}[t!]
    \centering 
    \includegraphics[width=1.0\linewidth]{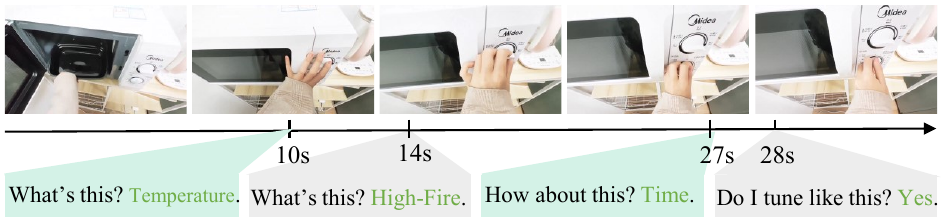}
    \caption{Example from \dataset~about a blind user demonstrating egocentric visual assistance. As she places her hands on various microwave dials, she asks a series of questions about what the dial controls, its position and settings and how to adjust it. 
    }
    \label{fig:intro_1}
    \vspace{-0.4cm}
\end{figure}

Our focus on blind users' needs for egocentric visual assistance allows us to explore several key challenges in video-language learning, including egocentric dynamic scene understanding with poor visual quality (\eg, unstable motion, object blur, and occlusions), in-situation user intention reasoning, assistive and reliable answer generation. Take Figure \ref{fig:intro_1} as an example; engaging in the user's activity from a first-person perspective is crucial to understanding and answering the questions. This specification requires capturing the gaze area, hand motions and reasoning about related visual content spatio-temporally. Finally, this should be achieved with poor quality visual inputs, as the regions of interest may be out of focus or occluded (\eg, the control panel in Figure \ref{fig:intro_1}).

With \dataset, we benchmark 16 recent MLLMs, covering the advanced open-source models (\eg, InternVL2.5 \cite{chen2024far}, Qwen2.5-VL \cite{wang2024qwen2}) and the closed-source ones (\eg, GPT-4o \cite{hurst2024gpt} and Gemini \cite{team2023gemini,comanici2025gemini}). For open-source models, we 
consider models that 1) achieve the state-of-the-art (SOTA) 
on common video QA benchmarks~\cite{yu2019activitynet,xiao2021next,maaz2024video,fu2024video}, and egocentric understanding benchmarks~\cite{mangalam2023egoschema,cheng2024videgothink,ye2024mm}. Additionally, we include 3 models (Video-LLaVA \cite{lin2023video} and LLaMA-VID \cite{li2024llama}, VILA1.5 \cite{lin2024vila}) that achieve SOTA 
on the blind image QA dataset VizWiz \cite{gurari2018vizwiz}.

Our experimental results show that 
all models struggle on \dataset. The best performer (GPT-4o) achieves an accuracy of 59.3\% and falls behind human performance of 87.4\% by a whopping $\sim$28\%. Interestingly, close-sourced models (\eg, Gemini 1.5 and 2.0), which often surpass open-source models in general-purpose VQA \cite{xiao2021next,mangalam2023egoschema,fu2024video} perform worse than top-performing open-source models (\eg, InternVL2.5). 
Also, models superior on egocentric VQA are not necessarily the best-performing. Our further analyses and investigations lead to the following primary observations:
\begin{itemize}[leftmargin=*]
     \item No single model wins across all assistance types, and most models are poor in egocentric navigation, safety warning, and communication, indicating research deficiency in these areas. 
    \item Models may correctly answer questions about the visual scene, though the answers may fail to meet users' needs for assistance, indicating a weakness in reasoning about user intentions. 
    \item  Models struggle in reasoning the change of spatial orientation relative to the users from the sequence of egocentric visual inputs.
    \item Models, especially the open-source ones, are sycophantic~\cite{zhao2024towards}; they hallucinate wrong and potentially malicious answers when the users' questions deviate the visual facts due to blindness.
    \item Finetuning with EgoBlind training data effectively benefits model performances, though a large gap remains compared to human capabilities.
\end{itemize}

Our work tackles significant challenges to construct the first VideoQA dataset (\dataset) to benchmark and promote research toward egocentric visual assistance for the blind.  We have comprehensively analyzed the behaviors of the leading MLLMs, revealing limitations and areas for improvements.  With these efforts, we hope that EgoBlind will serve as a foundation
for developing effective AI assistants to enhance the independence of the blind
and visually impaired.

\section{Related Work}
\label{sec:related}

\textbf{VQA for the Blind.}
Visual question answering (VQA) is the task of answering user questions about images and videos~\cite{antol2015vqa,xiao2024videoqa,yang2024robust}. 
One promising application area is to support the visually impaired.
Yet, the majority of advancements have been made in general-purpose settings~\cite{goyal2017making,xu2017video,xiao2021next,yu2019activitynet} 
with images and videos captured from third-person perspectives. The closest in aim is VizWiz \cite{gurari2018vizwiz}, which collects data from real human-powered VQA systems for helping blind users (\eg, BeMyEye\footnote{http://www.bemyeyes.org/}). 
VizWiz opens up the potential for visual assistance of the blind, but is limited in handling static images and object-centric questions; it fails to cater to the real-time and broader assistive needs of blind users in exploring the dynamic surroundings. Recent work VIEW-QA \cite{song2024video} 
captures the daily challenges faced by visually impaired individuals via using 360-degree egocentric wearable cameras. However, its videos and questions are collected 
from seeing actors who simulate the experiences of the blind, and thus may not reflect the daily lives and true needs of the blind for visual assistance, especially for those who have been blind for a long time.

\textbf{Egocentric VQA.}
Egocentric VQA has gained interest for its 
application value towards embodied assistance \cite{plizzari2024outlook,xiao2024videoqa}. 
Early effort EgoVQA \cite{fan2019egovqa} is small scale and limits its questions in challenging action recognition from first-person perspective with the absence of the camera wearer in the footage. Subsequent advancement EgoTaskQA \cite{jia2022egotaskqa} uses machine-generated questions to evaluate models’ task understanding capabilities. AssistQ \cite{wong2022assistq} is close to our aim for embodied assistance, but it limits to instructional videos for demonstration of tool use.
Ever since the release of Ego4D \cite{grauman2022ego4d}, lots of real-world egocentric QA tasks are explored \cite{lin2022egocentric,pramanick2023egovlpv2,Baermann_2022_CVPR,mangalam2023egoschema,di2024grounded,Cheng_2024_CVPR,cheng2024videgothink,ye2024mm}. However, they mostly aim at general-purpose egocentric visual understanding, or focus on a specific aspect of assistance, \eg, QAEgo4D \cite{Baermann_2022_CVPR} for episodic memory, EgoTextVQA \cite{zhou2025egotextvqa} for scene text understanding and EgoLifeQA \cite{yang2025egolife} for long context life assistants. To our best knowledge, there is so far no existing egocentric VQA dataset specially collected from real blind individuals.

\textbf{MLLMs for VQA.} 
By integrating visual information into powerful LLMs through lightweight connection modules (\eg, MLP \cite{liu2024improved}), MLLMs extend the capabilities of LLMs to converse with images and videos naturally like humans. MLLMs have significantly improved the landscape of VQA. This brings breakthrough over traditional VQA which answers questions by either multi-choice selection \cite{jang2017tgif,xiao2021next} or close-vocabulary classification \cite{xu2017video,yu2019activitynet}. While most MLLMs are developed for VQA from third-person views~\cite{xiao2024videoqa,maaz2024video,zhang2023video,lin2023video,xiao2024can,li2024llama,cheng2024videollama,wang2024qwen2,zhang2025videollama}, recent advancements~\cite{lin2022egocentric,pramanick2023egovlpv2,zhao2023learning,chen2024sharegpt4video,chen2024far,ye2024mm} specifically add egocentric videos for understanding. Nonetheless, all of they target a general-purpose egocentric visual understanding. In this paper, we will comprehensively examine their 
capabilities towards egocentric visual assistance for the blind users.

\section{\dataset~Dataset}
\label{sec:dataset}

\begin{wrapfigure}[10]{r}{0.6\textwidth}
    \centering
    \vspace{-0.5cm}
    \includegraphics[width=1.0\linewidth]{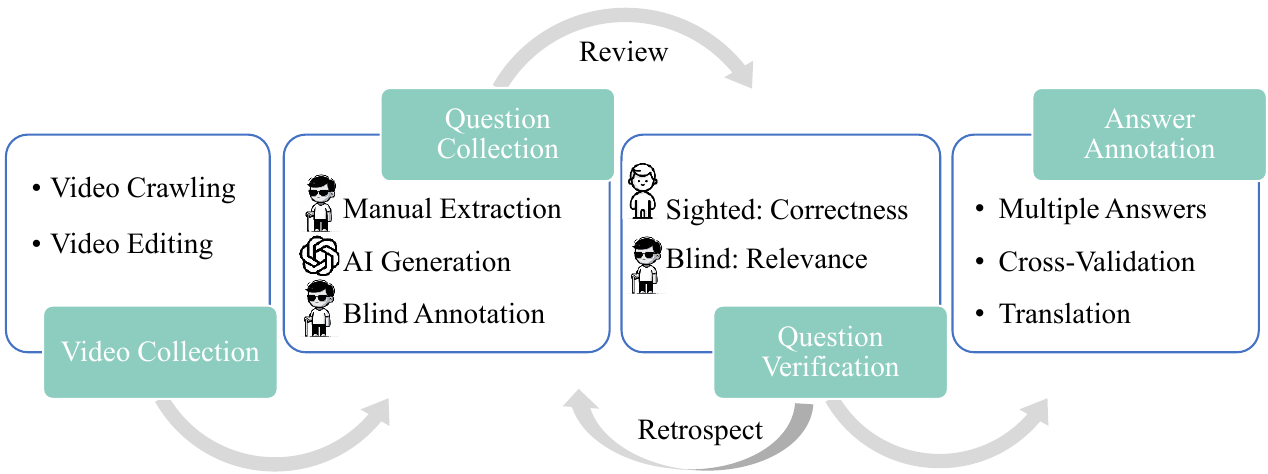}
    \vspace{-0.5cm}
    \caption{Data annotation pipeline.}
    \label{fig:data_collect}
\end{wrapfigure}
\subsection{Video Collection} 
We collect videos from video sharing platforms such as Bilibili and TikTok and design an annotation pipeline as illustrated in Figure \ref{fig:data_collect}. 
Specifically, we download 478 long-form egocentric videos (in batches)  of blind and visually impaired content creators.
These videos, captured using GoPro or mobile phones, document their daily lives while traveling, cooking, navigating through town, shopping, and in social gatherings. 
It provides near-real-world data on how visually impaired individuals perform daily tasks and solve problems in dynamic environments.
Since a single long video often contains an array of visual content, \eg, a summarized record of a visually impaired creator's week, we manually partition the videos into distinct segments based on the source edits. We exclude video moments with sharp scene transitions or strong staging and production indications (\eg, large subtitles, visual special effects, etc.) to 
focus primarily on life-logging. After partioning and filtering, we obtain 1,329 video clips with an average duration of 40 seconds.

\subsection{Question Collection} 
We obtain the questions via three different approaches:
1) \textbf{Manual Extraction}: We extract visually-assistive questions posed by the blind content creators in their videos. 
2) \textbf{AI-Generation}: We prompt GPT-4o \cite{hurst2024gpt} to act as a blind user and generate questions by engaging in their egocentric perspective. The generated questions are further verified and edited by both seeing and blind people, ensuring their correctness and alignment with blind users' true needs.
3) \textbf{Blind User Annotation}: For some videos, we describe the contents to blind annotators and ask them pose assistive questions as if they were in such a setting. The actual annotation is done together with the verification stage for AI generated questions. Our collection details are as follows:

\textbf{Manual Extraction} 
We first manually watch the video clips and extract the vision-assistive questions from the camera wearers, the timestamps of the questions are also recorded. It is worth noting that these in-video questions are answered by the blind users' sighted partners or others in the videos. Thus, to prevent answer leakage and ensure real-time QA, models can only access visual content up to the question timestamps to answer a specific question. We obtain a total of 541 questions during this process since qualified questions are sparse in the videos.

\textbf{AI-Generation.} To enrich the annotations, we consider generating questions by prompting GPT-4o {\cite{hurst2024gpt}} to act as blind questioners, followed by rigorous verification with real blind individuals.  We categorize the assistance scenarios into six groups based on our observation of video contents and the extracted questions: 
\emph{Information Reading}, \emph{Safety Warning}, \emph{Navigation}, \emph{Social Communication}, \emph{Tool Use}, and \emph{Other Resources}. 
Typical questions from each group is listed in Table \ref{tab:qa_examples}.
To ensure a diverse set of examples for each assistance type, the generation is conducted categorically with tailored prompts for each question type. In addition, we also prompt GPT to generate a reference answer for each question.

In our prompts, the generated QA pairs 
are ensured to be context-aware and aligned with the needs of the blind:
1) \textbf{Ego-centric}: All questions should be framed from the perspective of the visually impaired individual (camera wearers). 
2) \textbf{Practical}: The questions must reflect practical needs of the blind in the given situation.
3) \textbf{Video-Level Questions}: Questions should reflect temporal events in the video, emphasizing dynamic characteristics and requiring multiple frames to answer. 
4) \textbf{Online Context}: Questions must pertain to the online context of the video and exclude any external information beyond the visible content up to the question time stamp. 
Other details are presented in the Appendix \ref{sec:aigen}.
\renewcommand{\arraystretch}{0.5}
\begin{table}[t!]
    \centering
    \caption{QA examples categorized by different assistance needs.}
    \resizebox{1.0\linewidth}{!}{
    \begin{tabular}{p{2cm} c p{10cm}}
        \toprule
        \textbf{Category} & \textbf{\#Q / \#V} & \textbf{QA Examples (Multiple reference answers)} \\
        \midrule \midrule
        Information Reading & 2,464 / 1,040 & 
        Q: What floor is the elevator currently on?\newline
        A: 1st floor / At this time on the first floor. \\
        \midrule
        Safety\newline Warnings & 1,260 / 686 & 
        Q: Is it safe to cross the street now? \newline
        A: No. / No, there are cars on both sides. / A car is about to pass through the street, it is recommended to wait a little. \\
        \midrule
        Navigation & 751 / 438 & 
        Q: Where is the entrance to the building? \newline
        A: Just a few steps ahead of you. / Directly in front. \\
        \midrule
        Social\newline Communication & 153 / 131 & 
        Q: Who is the person talking to me? \newline
        A: He is a delivery guy. / Delivery man. \\
        \midrule
        Tool Use & 288 / 197 & 
        Q: How do I turn on the stove? \newline
        A: Turn the knob on the front of the stove clockwise until you feel a click. / By clock in the switch button on the stove. / Rotate the button below. / Turn the switch to the left. \\
        \midrule
        Other \newline Resources & 395 / 280 & 
        Q: Is there anyone nearby that I can ask for directions? \newline
        A: Yes. / Yes, there is a person sitting in the front. / Yes, there is a security guard ahead. / Yes, there is front-desk security. \\
        \bottomrule
    \end{tabular}}
    \label{tab:qa_examples}
    \vspace{-0.4cm}
\end{table}

\subsection{Manual Verification and Answer Annotation} 
Our \textbf{verification} is done in three stages involving both seeing and visually impaired people. The first three authors are engaged in all stages for quality control. Concretely, \textbf{in the first stage}, the authors check the QA quality after generating a limited number of questions for each assistance type. Related issues are recorded and reflected in the updated prompts for subsequent generations to improve the generation quality. In such an alternate way, we generate 16,560 questions in batches. 

\textbf{In the second stage}, we invite 63 volunteers seeing individuals to thoroughly review the questions. In this process, we remove questions that are 1) redundant in meaning; 2) vague and difficult to answer directly (\eg, ``How do I use my phone?''); and 3) meaningless (\eg, ``How do I use my white cane?''). We also remove a moderate number of 
questions that are 1) not related to visual inputs (\eg, ``How heavy is this box?'') and 2) not engaged in first-person perspective (\eg, ``what is the man (camera wearer) doing?''). Note that all ambiguous questions are further cross-validated by the three authors to either remove or edit them. After this stage, 5,080 questions remained; more than 80\% of them have been human-edited by this point. Additionally, the 541 extracted questions also undergo a check for language translation errors, resulting in a total of 5,621 valid questions after this stage.

\textbf{In the third stage}, we sample 30\% (1.5K) of the questions and invite 111 blind volunteers to review them for relevance to blind users' needs for visual assistance. The sampling are visual situation based to cover all visual situations in EgoBlind (see details in Appendix \ref{sec:blindstudy}). 
Specifically, the blind volunteers are requested to score (from the lowest 0 to the highest 5) the confidence that a question would be asked under a specific visual scenario. Each question is scored by 10 to 33 volunteers. Additionally, the bind volunteers are encouraged to pose additional new questions. 
Afterwards, we remove the questions with an average confidence score lower than 1, and add approximately 300 new questions directly from the blind volunteers (\textbf{Blind User Annotation}). 
The result observations (what question to be deleted or added) are then reflected throughout the whole dataset by the authors.
Finally, we obtain 5,311 valid questions. Their distributions are listed in Table \ref{tab:qa_examples}.

\textbf{Multiple Answer Annotation.} 
For each question, we annotate multiple ground-truth answers by inviting 21 university students to watch the related video. To enhance the persuasiveness of the standard answers and reduce subjectivity, each question is assigned to three or four annotators for answering. Only one may see the generated answer (for generated questions only) and must verify and edit it as needed. The others are required to directly answer the question for assistance purpose. 
If there is no generated answer, \eg, for the manually extracted or annotated question, the annotators need to answer the question themselves.
We resolve contradictory answers through consensus between the authors and the majority of annotators. Redundant answers are avoided by merging similar responses. Ultimately, each question has 1 to 4 valid answers (3 on average). Over 67\% of the generated answers are edited by human annotators, and most unedited answers are deterministic, such as yes/no and numbers. Examples of answers are shown in Table \ref{tab:qa_examples}.

\begin{figure}[t!]
  \centering
  \begin{subfigure}{0.25\linewidth}
    \includegraphics[width=1.0\linewidth,height=0.8\linewidth]{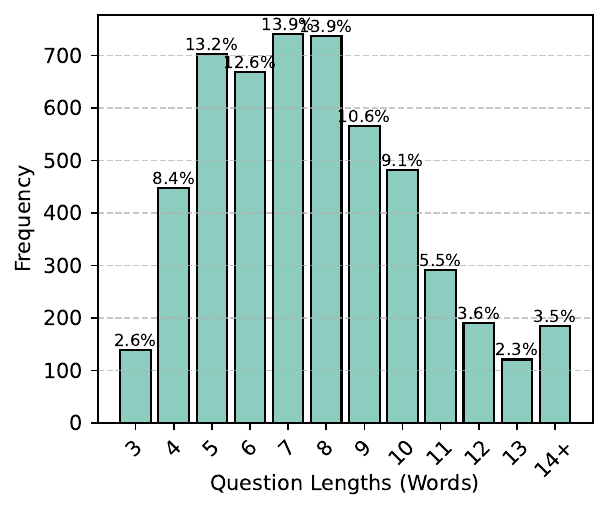}
    \vspace{-0.5cm}
    \caption{Question length.}
    \label{fig:short-a}
  \end{subfigure}
   \begin{subfigure}{0.23\linewidth}
    \includegraphics[width=1.0\linewidth, height=0.87\linewidth]{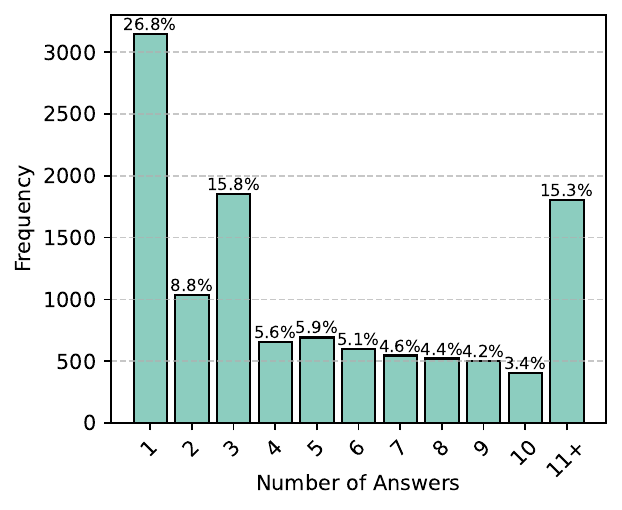}
    \vspace{-0.5cm}
    \caption{Answer length.}
    \label{fig:short-b}
  \end{subfigure}
  \begin{subfigure}{0.25\linewidth}
    \includegraphics[width=1.0\linewidth,height=0.8\linewidth]{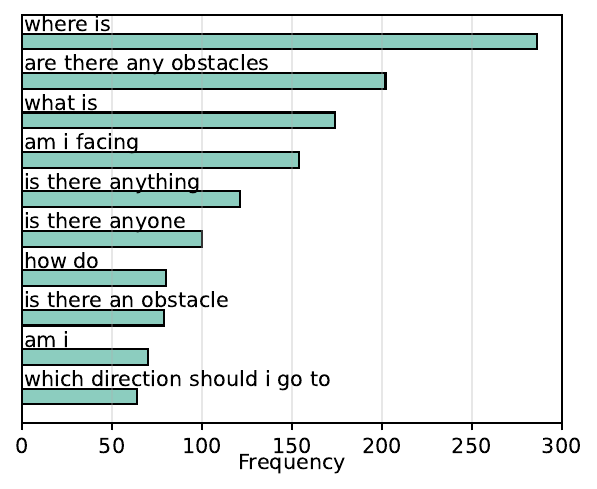}
    \vspace{-0.5cm}
    \caption{Top-10 questions.}
    \label{fig:short-c}
  \end{subfigure}
   \begin{subfigure}{0.25\linewidth}
    \includegraphics[width=1.0\linewidth,height=0.8\linewidth]{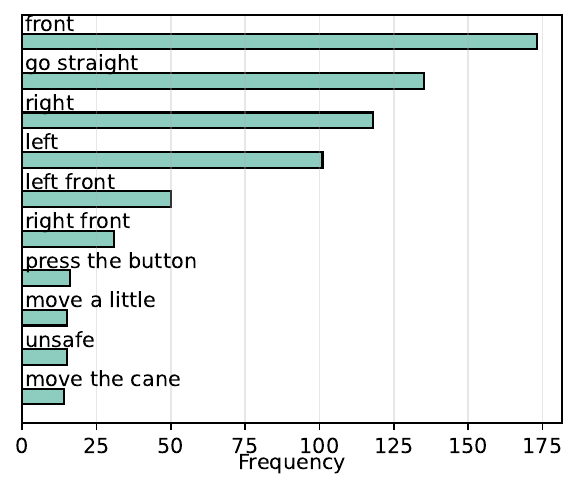}
    \vspace{-0.5cm}
    \caption{Top-10 answers*.}
    \label{fig:short-d}
  \end{subfigure}
  
  \begin{subfigure}{0.33\linewidth}
    \includegraphics[width=1.0\linewidth]{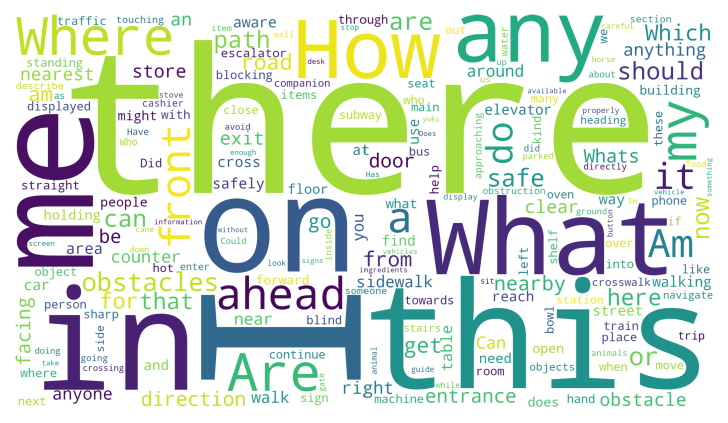}
    \caption{Word cloud of questions.}
    \label{fig:short-e}
  \end{subfigure}
  \begin{subfigure}{0.33\linewidth}
    \includegraphics[width=1.0\linewidth]{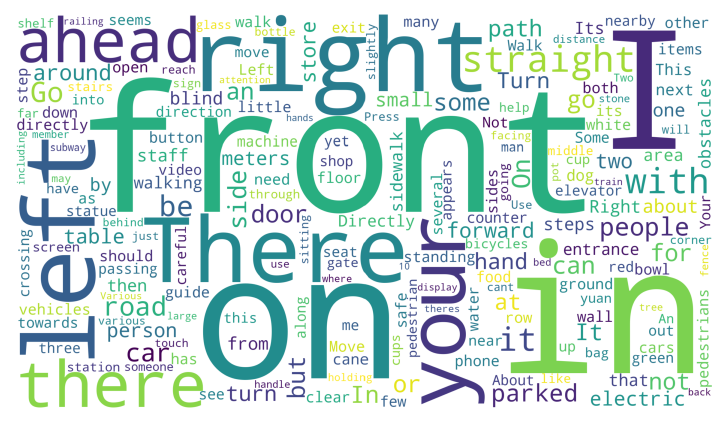}
    \caption{Word cloud of answers*.}
    \label{fig:short-f}
  \end{subfigure}
   \begin{subfigure}{0.32\linewidth}
    \includegraphics[width=1.0\linewidth, height=0.59\linewidth]{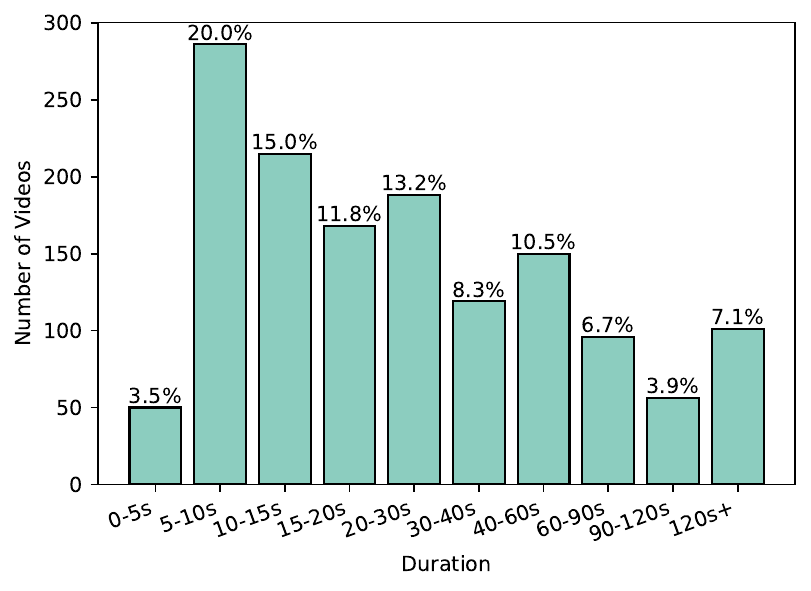}
    \caption{Distribution of video durations.}
    \label{fig:short-g}
  \end{subfigure}
  
  \caption{Statistic analysis of \dataset. *: We omit 1,123 (22.6\%) ``yes/no'' and 541 (11.0\%) ``do not know'' answers in (d) and (f) for better presentation. (Please zoom in for better view.)}
  \label{fig:qas_dis}
  \vspace{-0.6cm}
\end{figure}

\subsection{Statistic Analysis}
\begin{wraptable}[6]{r}{0.47\textwidth}
    \centering
    \small
    \vspace{-0.36cm}
    \caption{ Statistics of the \dataset~datasets}
    \begin{tabular}{ccc|ccc}
    \toprule
    \multicolumn{3}{c}{\textbf{Videos}}  & \multicolumn{3}{c}{\textbf{Questions}} \\
    \midrule
    Train & Test & All  & Train & Test & All \\
    \midrule
    673 & 656 & 1,329  & 2,746 & 2,565 & 5,311 \\
    \midrule
    \end{tabular}
\label{tab:statistics_examples}
\vspace{-0.3cm}
\end{wraptable}
\textbf{Data Split.} The data collection and annotation process takes about 10 months. We finalize 5,311 questions across 1,329 videos. For effective and efficient evaluation, we split half of the videos (656) along with their QA pairs (2,565) as a test set, while leaving the remaining for instruction tuning towards model development. 
The split ensures that video clips from the same source video do not appear in both sets (we also study a user-specific data-split strategy in Appendix \ref{sec:inv}). 
Additionally, we use Gemini 2.0 to probe if there is a significant discrepancy of model behaviors between the test set and the whole dataset. We find that the prediction accuracy differences are quite small, ranging from 0.2\% to 0.5\% across different question types. This suggests that the test set is sufficient for reliable evaluation.  
The detailed numbers are presented in Table \ref{tab:statistics_examples}.

\noindent\textbf{Questions.} 
Figure \ref{fig:short-a} shows that more than 80\% of the questions are shorter than 10 words, with an average length of 7.9 words. The relatively short questions could be attributed to the use of more spoken language in practice. The word cloud in Figure \ref{fig:short-e} shows that the questions are from blind users' first-person perspective (``I'', ``me'', ``my'') and feature referential concepts, such as ``this'',  ``there'', ``that'', ``it'' and ``now'', which requires context-specific interpretation. 
Figure \ref{fig:short-c} shows that blind people are often interested in locating something (``where is'') and checking for potential safety issues (``are there any obstacles'').

\noindent\textbf{Answers.} 
Figure \ref{fig:short-b} shows that the answers have an average length of 5.7 words. Near half of the answers have more than 3 words, with 19.4\% over 10 words. A detailed study shows that the answers for tool use are significantly longer and more complex than other assistive questions; the average answer length is 12.1 words for tool use and 4 to 6 words for others. The most frequent answers are ``yes'' (13.0\%), followed by ``I do not know'' (10.7\%) and ``no'' (9.6\%). The answer of ``I do not know'' is due to poor visual quality or the answer not framed by the blind users.  Additionally, it may be attributed to our online QA task setting, \ie, the answer is not visible up to the question moment. We keep these unanswerable questions to evaluate if models can reject to answer rather than hallucinate potentially malicious answers. 
For better analysis, we omit these answers and analyze the remaining frequent answers and answer word clouds in Figure \ref{fig:short-d} and \ref{fig:short-f}, respectively. We find that the answers are mostly about directions, navigation and locations. It is worth mentioning that the locations and directions are often relative to the camera wearers (questioners). 

\noindent\textbf{Videos.} 
Figure \ref{fig:short-g} shows that the vast majority of the videos (82.3\%) are within 1 minute, with 17.7\% exceeding 1 minute and 7.1\% running longer than 2 minutes. The mean video duration is 40.0 seconds, reflecting the average duration each time a visually-impaired person recording their daily moment. Additionally, we conduct a comparison between the egocentric video content captured by sighted people (\eg, Ego4D \cite{grauman2022ego4d} videos) and those are blind (examples are presented in Appendix \ref{sec:compvideo}). Differences arise from 1) \textbf{Composition and Focus:} Blind individuals often produce footage where subjects are off-center or out of focus (refer to Figure \ref{fig:intro_1} and Appendix \ref{sec:compvideo}). 2) \textbf{Camera Orientation and Stability:} Blind individuals struggle to maintain consistent camera orientation, leading to  potentially tilted or unstable footage. 3) \textbf{Environmental Awareness:} Blind individuals inadvertently capture obstructed views or poorly lit scenes. Nevertheless, the video resolution is higher, possibly because the videos are crawled from content creators on video sharing platforms.

\begin{table}[t!]
\centering
\small
\caption{Ego-VQA dataset comparison. BV: Video captured by the blind. BQ: Question posed or verified by the blind. RT: Answer based on the video content up to the question timestamp. MA: Multiple answers for each question. QC: Question Category. OE/MC: Open-Ended/Multi-Choice.
}
\begin{threeparttable}
\begin{tabular}{llccccccccccc}
\toprule
\multicolumn{2}{c}{\textbf{Datasets}} & \textbf{\# V} & \textbf{\# Q} & {\bf VLen(s)} & \textbf{BV} & \textbf{BQ} & \textbf{RT} & \textbf{MA} & \textbf{QC}  & \bf Task \\ 
\midrule
\multirow{4}{*}{\makecell{General \\ Purpose}}
& EgoTaskQA~\cite{jia2022egotaskqa} & 2K & 40K & 25.0 & \xmark & \xmark  & \xmark & \xmark & \cmark & OE  \\
& VidEgoThink~\cite{cheng2024videgothink} & 217 & 600 & 23.4 &  \xmark & \xmark  & \xmark & \xmark & \cmark & OE  \\
& EgoSchema~\cite{mangalam2023egoschema} & 5K & 5K & 180.0 & \xmark & \xmark  & \xmark  &  \xmark  &  \xmark & MC \\  
& EgoMemoria~\cite{ye2024mm} & 629 & 7K & 858.5 & \xmark & \xmark & \xmark & \xmark & \xmark  & MC \\ 
\midrule
\multirow{4}{*}{Assist}
& QAEgo4D~\cite{Baermann_2022_CVPR} & 1.3K & 14.5K & 495.1 & \xmark & \xmark &  \xmark  & \xmark  & \xmark  & OE \\ 
& AssistQ~\cite{wong2022assistq}  & 100 & 531 & 115.0 & \xmark & \xmark & \xmark &  \xmark & \cmark   & MC \\ 
& EgoTextVQA~\cite{zhou2025egotextvqa} & 1.5K & 7.0K & 101.7& \xmark & \xmark & \cmark & \xmark & \cmark  & OE \\ 
\midrule
\multirow{3}{*}{Blind}
& VizWiz~\cite{gurari2018vizwiz} & - & - & - & \cmark & \cmark & \xmark & \cmark & \xmark  & OE \\ 
& VIEW-QA~\cite{song2024video}  & 1.0K & 4.1K & 34.4 &  \xmark  & \xmark & \xmark & \xmark & \cmark  & OE \\ 
\rowcolor{lightgray} 
& \dataset~(Ours)  & 1.3K & 5.3K & 40.0 & \cmark &\cmark & \cmark & \cmark & \cmark & OE \\ 
\bottomrule
\end{tabular}
\end{threeparttable}
\label{tab:dataset_compare}
\vspace{-4mm}
\end{table}
\subsection{Dataset Comparison}  
Unlike most Ego-VQA datasets (the first block of Table \ref{tab:dataset_compare}) which focus on general-purpose visual understanding, \dataset~emphasizes assisting blind individuals.
Compared to other assist-oriented datasets (the middle block), \dataset~covers the wide aspects of visual assistance for the blind in daily life, versus episodic memory in QAEgo4D \cite{Baermann_2022_CVPR}, tool use in AssistQ \cite{wong2022assistq}, manipulation in HoloAssist \cite{wang2023holoassist} (proactive assistance), or scene-text reading in EgoTextVQA \cite{zhou2025egotextvqa}.

Compared to the blind-related QA datasets (the bottom block), \dataset~advances VizWiz \cite{gurari2018vizwiz} by enabling egocentric live QA with more visual information about real-world dynamics and episodic memory support. 
Although the videos in the VIEW-QA \cite{song2024video} dataset offer 360-degree visual information, the videos and questions are simulated by seeing actors. 
In contrast, the videos in \dataset~are entirely filmed by blind or visually impaired individuals themselves, authentically reflecting their real-life scenarios. Moreover, the questions in \dataset~are posed or verified by blind individuals of different blind ages, further ensuring the dataset’s practicality and relevance. Finally, \dataset~supports an online QA setting with timestamp annotations for each question; answers are based on the video contents prior to the question timestamp.

\section{Experiment}
\label{sec:experiment}

\textbf{Evaluation.} Following popular practices for LLMs~\cite{maaz2024video,song2024moviechat}, we use the GPT score as a metric for evaluating generated answers. Specifically, we prompt GPT-4o mini to assess the semantic similarity between a models’ prediction and the ground truth (reference answers), and answer with 'yes' if they are judged as the same. We then obtain the accuracy (0-100\%) as the percentage of ‘yes’ answers evaluated.
Noteworthy, for each question, a correct prediction is identified if it achieves \emph{yes}-response with any one of the reference answers. Moreover, the evaluation prompts are manually finetuned to reach a maximal agreement (0.88) between human and AI reviewers (details in Appendix \ref{sec:appeval}).

\textbf{Model Setup.} To benchmark the challenges carried by \dataset, we comprehensively analyze 16 contemporary MLLMs, including 12 open-source models and 4 closed-source ones (via APIs). The choice of the open-source models are based on the following criteria: 1) Achieve SOTA results on common video QA benchmarks such as NExT-QA \cite{xiao2021next}, VideoChatGPT-Bench \cite{maaz2024video} and Video-MME \cite{fu2024video}. Corresponding models are LLaVA-OV \cite{li2024llava}, CogVLM2-Video \cite{hong2024cogvlm2}, Video-LLaMA3 \cite{zhang2025videollama}, InternVL2.5 \cite{chen2024far}; 2) Achieve SOTA results on general-purpose egocentric VideoQA benchmarks such as EgoSchema \cite{mangalam2023egoschema}, VidEgoThink \cite{cheng2024videgothink} and EgoMemoria \cite{ye2024mm}. Corresponding models are LLaVA-Video \cite{zhang2024video}, MiniCPM-V 2.6 \cite{yao2024minicpm} and Qwen2.5-VL\cite{wang2024qwen2}; 3) Achieve SOTA on the image blind QA dataset VizWiz \cite{gurari2018vizwiz}, such as Video-LLaVA \cite{maaz2024video}, LLaMA-VID\cite {li2024llama} and VILA 1.5 \cite{lin2024vila}. Notably, for each question, we uniformly sample the video content up to the question timestamp for answer prediction, thus to match the live QA task setting. Moreover, we design customized prompts tailored for answering blind users' questions for better performance. Additionally, we obtain human performance by inviting 3 university students who did not participant in annotation.

\begin{table*}[t!]
\centering
\caption{QA accuracy of different models on \dataset.}
\resizebox{\textwidth}{!}{ 
\begin{threeparttable}
\begin{tabular}{lccccccccc|c}
\toprule
\textbf{Methods} & \textbf{LLM} & \textbf{Size} & \textbf{\#F} & \textbf{Tool} & \textbf{Info.} & \textbf{Navi.} & \textbf{Safe} & \textbf{Com.} & \textbf{Res.} & \textbf{Overall} \\
\midrule
\rowcolor{lightgray}
\textbf{Human} & - & - & - & 70.4  &  87.0  &  83.1  &  91.9   &94.7  & 96.6   & 87.4  \\
\midrule
\multicolumn{10}{l}{\textit{Open-source Models}} \\
\midrule
ShareGPT4Video \cite{chen2024sharegpt4video}  &  LLaMA3-8B & ori & 16 &  25.5  & 32.6   &  20.7 &  43.3  &  38.9   & 28.3 & 32.9 \\
CogVLM2-Video \cite{hong2024cogvlm2} & LLaMA3-8B & 224\textsuperscript{2} & 24 & 32.2 & 44.5 & 14.0 & 52.7 & 43.1 & 32.4 & 40.3 \\
Video-LLaMA3 \cite{zhang2025videollama} & Qwen2.5-7B & ori & 1fps & 53.0 & 51.9 & 38.1 & 50.6 & 41.7  & 50.3 & 49.2 \\
InternVL2.5-8B \cite{chen2024far} & InternLM2\_5-7B & 448\textsuperscript{2} & 8 & 61.1 & 54.6  & 42.2 & 58.0 &  44.4 & 52.6 & 53.5 \\
LLaVA-OV \cite{li2024llava} & Qwen2-7B & 384\textsuperscript{2} & 16 & 61.1 & 56.4 & 29.5 & {\bf65.8} & {\bf58.3} & 50.9 & 54.5 \\
InternVL2.5-26B \cite{chen2024far} & InternLM2\_5-20B & 448\textsuperscript{2} & 8 & {\bf72.5} & \underline{56.9} & 47.4 & 54.1 & 43.1 & \underline{53.2} & 55.0 \\
\midrule
MiniCPM-V 2.6 \cite{yao2024minicpm} & Qwen2-7B & 384\textsuperscript{2} & 1fps & 53.7 & 46.5 & 37.8 & 28.9 & 37.5 & 41.0 & 40.7 \\
Qwen2.5-VL \cite{wang2024qwen2} & Qwen2.5-7B & ori & 1fps &  51.0 & 50.1 & 28.2 & 48.5 &  43.1 & 38.2 & 45.5 \\
LLaVA-Video \cite{zhang2024video}  & Qwen2-7B & 384\textsuperscript{2} & 1fps & 44.3 & 53.4 & 32.6 & \underline{62.0} & \underline{50.0} & 49.7 & 51.5 \\
\midrule
Video-LLaVA \cite{maaz2024video} &Vicuna-7B & 224\textsuperscript{2} & 8 & 22.8 & 41.2 & 21.2 & 47.2 & 38.9 & 35.3 & 38.1 \\
LLaMA-VID \cite{li2024llama} & Vicuna-7B& 224\textsuperscript{2} & 1fps & 32.2 & 40.5 & 20.7 & 49.4 & 36.1 & 41.6 & 39.1 \\
VILA1.5 \cite{lin2024vila} & LLaMA3-8B & 336\textsuperscript{2} & 8 & 49.7 & 50.5 & 25.9 & 60.6 &  47.2  & 41.0  &  48.2 \\
\midrule
\multicolumn{10}{l}{\textit{Closed-source Models}} \\
\midrule
Gemini 2.0 Flash  & - & ori & 32 & 61.1  & 54.5 & \underline{50.5} &  39.1 & 47.2& 49.1 &  49.9  \\
Gemini 1.5 Flash & - & ori & 32 & \textbf{72.5} &  54.4  &  43.5  &  50.6 & 38.9 & 45.7 & 51.8 \\
Gemini 2.5 Flash  & - & ori & 32 & \underline{67.1}  & 57.6 & 47.7  &  57.8 & 47.2 & 50.3 &  \underline{56.0}  \\
GPT-4o  & - & ori & 32 &  66.4  &  {\bf61.2} & {\bf52.6} & 58.8 & 47.2 & {\bf62.4} & {\bf59.3} \\
\bottomrule
\end{tabular}
\end{threeparttable}
}
\label{tab:comparison}
\vspace{-0.3cm}
\end{table*}

\subsection{Benchmarking Analysis}

Table \ref{tab:comparison} presents the performances of different models. We summarize the following observations:
    \textbf{(1)} None of the models achieves the desired level of performance on \dataset, all lagging behind human performance by a whopping 54\%$\sim$28\%, suggesting significant room for improvements.\\
    \textbf{(2)} The models that are superior at general-purpose egocentric VQA (\eg, LLaVA-Video) and image blind-VQA (\eg, VILA1.5) are not the best-performing, depicting unique challenges of EgoBlind.\\
    \textbf{(3)} No single model wins across all question types. Answering ``Navigation'' questions is the most challenging task for almost all models, indicating a significant limitation of MLLMs in this field. \\  
    \textbf{(4)} While most other models struggle in answering questions about tool use, Gemini 1.5 even surpasses human performance, demonstrating its rich knowledge outweighing human individuals.\\
    \textbf{(5)} Stronger LLMs and larger visual resolution often bring better performance, while more frames do not always help (\eg, 8 frames are enough for InternVL to surpass other open-source models.).

\subsection{Assistance-Related Challenges}
We further reveal some specific challenges pertaining to egocentric visual assistance for the blind by analyzing the common failure cases.

\begin{figure}[t!] 
    \centering
     \includegraphics[width=1.0\linewidth]{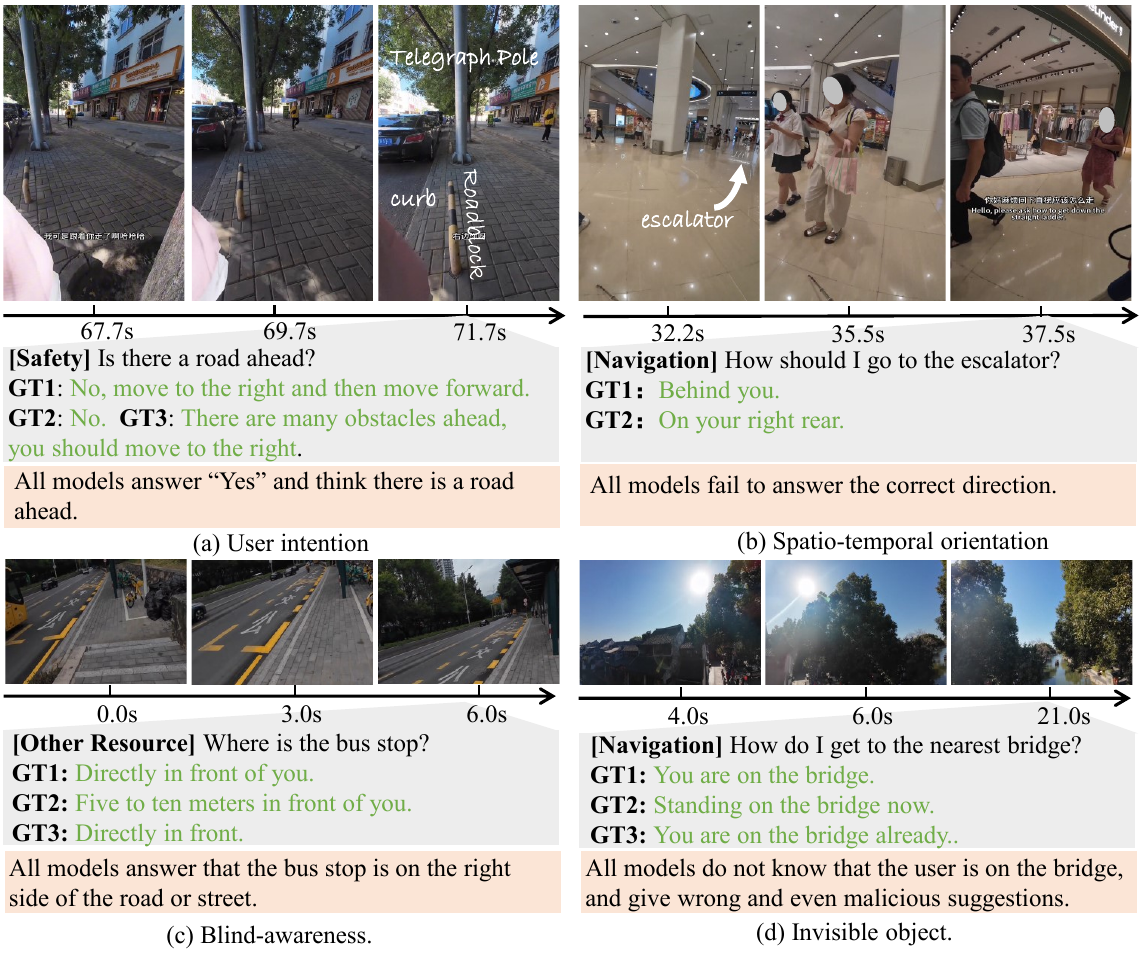} 
      \vspace{-0.5cm}
    \caption{Common failure cases of tested MLLMs. The models fail to (a) reason user intention, (b) understand real-time spatial orientation, provide (c) assistive and (d) reliable answers.} 
    \label{fig:visres}
    \vspace{-0.3cm}
\end{figure}

\textbf{User Intention.} Reasoning about user intentions behind the questions is key for effective assistance. Yet, all models fall short in such a capability. For example, in Figure \ref{fig:visres}(a) and (c), all models fail to generate helpful answers for the blind, though the answers are objectively correct to the visual contents. It is worth mentioning that this is despite explicitly prompting the models to answer questions of blind individuals to provide visual assistance. 

\textbf{Spatial Orientation Change.} MLLMs exhibit significant shortcomings in weaving together the temporal frames to reason the spatial orientations relative to the users' real-time location. A typical example is shown in Figure \ref{fig:visres}(b), where indoor navigation 
is requested by the blind user. The escalator was framed prior to the question moment, and the models have to retrieve the related moment and reason the users' orientation change after that moment, which shows extreme challenge to all MLLMs.

\textbf{Reliable Answers.} Providing reliable answers is of crucial importance in egocentric visual assistance. However, all MLLMs tend to be sycophantic when the users ask something that deviates the visual facts due to blindness. Take Figure \ref{fig:visres}(d) as an example, the blind user is already on the bridge but ask for navigating it. All models fail to remind the user that he is on the bridge, but either answer ``I do not know'', or give wrong or even malicious suggestions, such as ``take the boat'', ``walk straight''.

We also analyze other challenges related to streaming VQA, scene text recognition, and referential words in spoken languages in Appendix \ref{sec:casestudy}.

\begin{figure*}[tb]
\begin{minipage}{\textwidth}
\begin{minipage}[t]{0.33\textwidth}
\makeatletter\def\@captype{table}\makeatother\caption{Single frame inputs.}\label{tab:sigframe}
\vspace{-0.3cm}
\begin{threeparttable}
\resizebox{1\textwidth}{!}{
    \setlength\tabcolsep{5.3pt}
        \renewcommand\arraystretch{1.045}
    \begin{tabular}{lccccc}
    \toprule
    Models & Acc. (bef) & Acc. (aft) \\ 
    \midrule
    InternVL2.5-26B & 55.0 & 53.4 \textcolor{red}{$\downarrow$ 1.6} \\
    VILA1.5 & 48.2 & 47.2 \textcolor{red}{$\downarrow$ 1.0} \\
    GPT-4o & 59.3 & 58.2 \textcolor{red}{$\downarrow$ 1.1}  \\
    \bottomrule
    \end{tabular}
}
\end{threeparttable}
\end{minipage}
\begin{minipage}[t]{0.33\textwidth}
\makeatletter\def\@captype{table}\makeatother\caption{Normal QA prompts.}\label{tab:blindpromt}
\vspace{-0.3cm}
\begin{threeparttable}
\resizebox{1\textwidth}{!}{
\setlength\tabcolsep{5.3pt}
\renewcommand\arraystretch{1.03}
\begin{tabular}{lccccc}
    \toprule
    Models & Acc. (bef) & Acc. (aft) \\ 
    \midrule
    InternVL2.5-26B &  55.0 & 54.0 \textcolor{red}{$\downarrow$ 1.0} \\
    Gemini 2.5 Flash & 56.0 & 55.2 \textcolor{red}{$\downarrow$ 0.8} \\
    GPT-4o & 59.3 & 58.1 \textcolor{red}{$\downarrow$ 1.2}  \\
    \bottomrule
\end{tabular}
}
\end{threeparttable}
\end{minipage}   
\begin{minipage}[t]{0.33\textwidth} 
\makeatletter\def\@captype{table}\makeatother\caption{Instruction Tuning.}\label{tab:itune}
\vspace{-0.3cm}
\begin{threeparttable}
\resizebox{1\textwidth}{!}{
\setlength\tabcolsep{5.3pt}
\renewcommand\arraystretch{1.045}
\begin{tabular}{lcccc}
    \toprule
    Models & Acc. (bef) & Acc. (aft)\\ 
    \midrule
    Qwen2.5-VL & 45.5 & 50.2 \textcolor{green}{$\uparrow$ 4.7} \\
    LLaVA-OV &  54.5 & 57.4 \textcolor{green}{$\uparrow$ 2.9} \\
    InternVL2.5-8B & 53.5 & 58.1 \textcolor{green}{$\uparrow$ 4.6} \\
    \bottomrule
\end{tabular}
}
\end{threeparttable}
\end{minipage}
\end{minipage}
\vspace{-0.2cm}
\end{figure*}

\begin{figure}[t!]
  \centering
  \begin{subfigure}{0.32\linewidth}
    \includegraphics[width=1.0\linewidth]{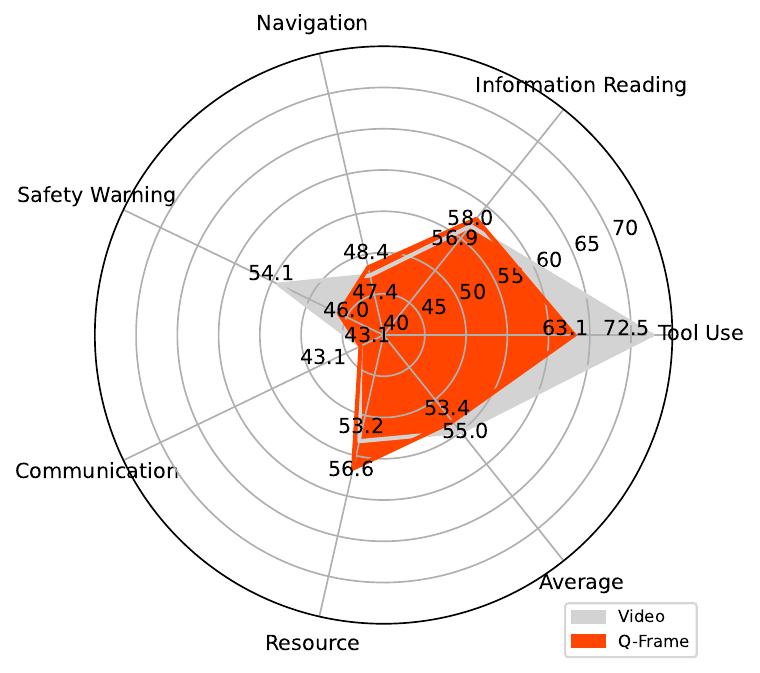}
    \caption{InternVL2.5-26B.}
    \label{fig:internvl}
  \end{subfigure}
  \begin{subfigure}{0.32\linewidth}
    \includegraphics[width=1.0\linewidth]{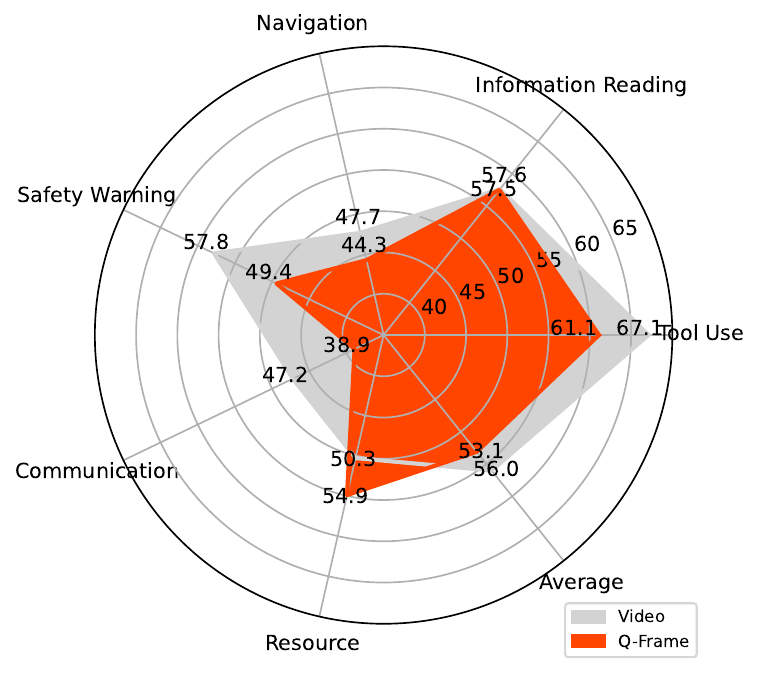}
    \caption{Gemini 2.5 Flash.}
    \label{fig:gemini}
  \end{subfigure}
   \begin{subfigure}{0.32\linewidth}
    \includegraphics[width=1.0\linewidth]{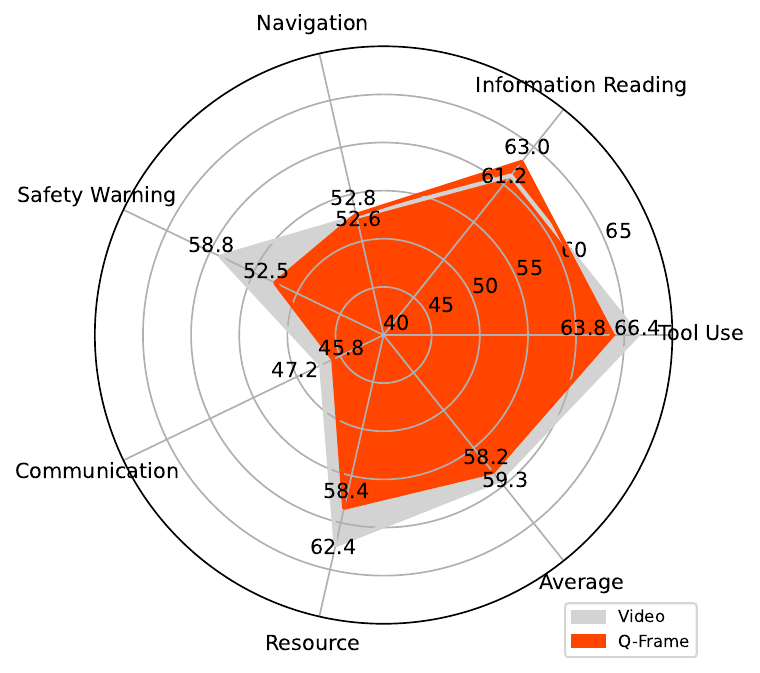}
    \caption{GPT-4o.}
    \label{fig:gpt4}
  \end{subfigure}
  \caption{Uniform sampling \vs single frame (Q-Frame) input at the question timestamp. The overall QA accuracy declines slightly when replacing video with a Q-Frame.}
  \label{fig:aba-vf}
\end{figure}

\begin{figure}[t!]
  \centering
  \begin{subfigure}{0.32\linewidth}
    \includegraphics[width=1.0\linewidth]{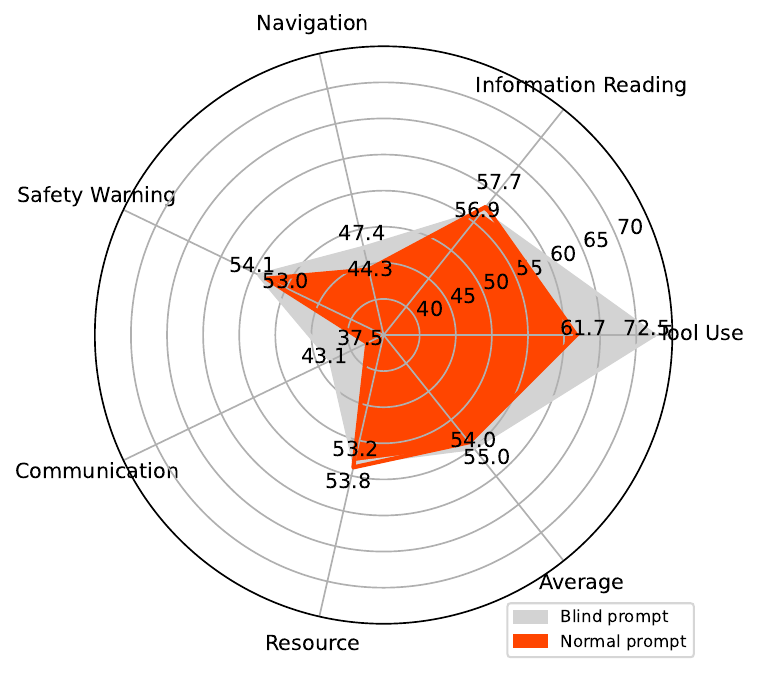}
    \caption{InternVL2.5-26B.}
    \label{fig:pt_internvl}
  \end{subfigure}
  \begin{subfigure}{0.32\linewidth}
    \includegraphics[width=1.0\linewidth]{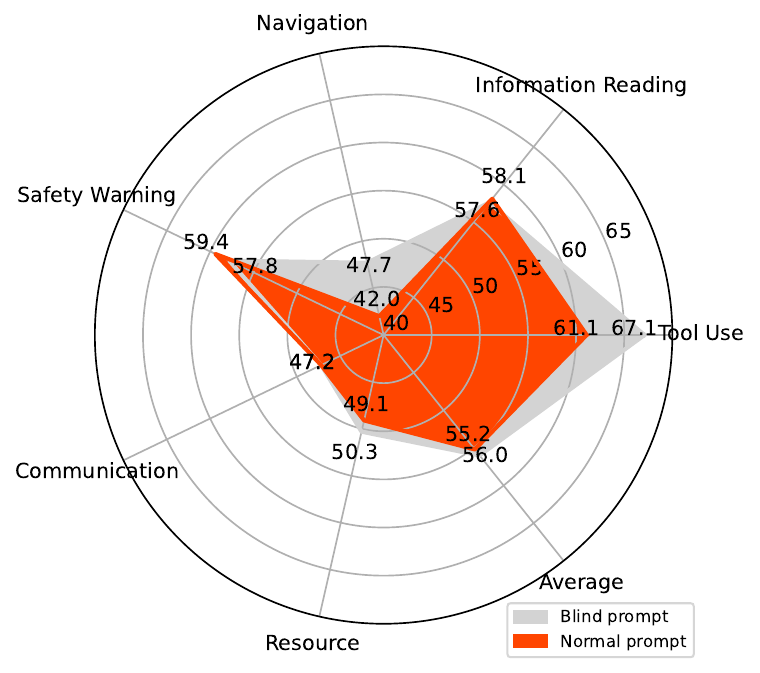}
    \caption{Gemini 2.5 Flash.}
    \label{fig:pt_gemini}
  \end{subfigure}
   \begin{subfigure}{0.32\linewidth}
    \includegraphics[width=1.0\linewidth]{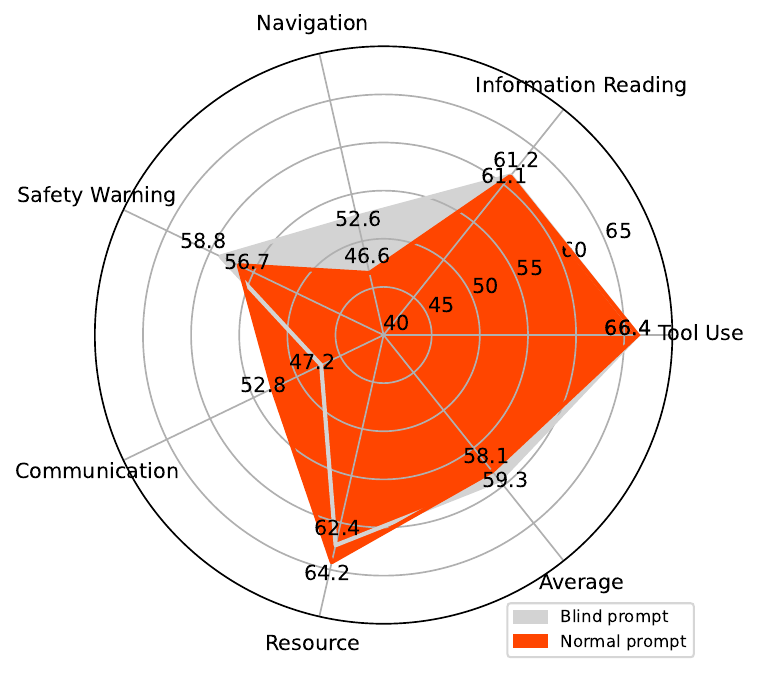}
    \caption{GPT-4o.}
    \label{fig:pt_gpt4}
  \end{subfigure}
  \caption{Blind aware prompt (Appendix Table \ref{tab:blind-prompt}) \vs Normal VQA prompt (Appendix Table \ref{tab:normal-prompt}). The overall accuracy declines without blind-specific prompting. }
  \label{fig:aba-prompt}
\end{figure}

\begin{figure}[t!]
  \centering
  \begin{subfigure}{0.32\linewidth}
    \includegraphics[width=1.0\linewidth]{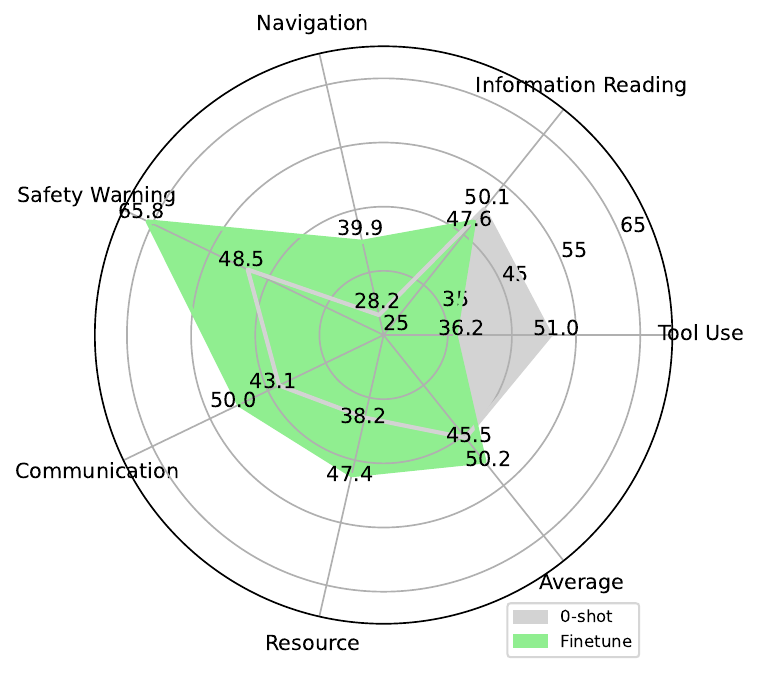}
    \caption{Qwen2.5-VL.}
    \label{fig:it_internvl}
  \end{subfigure}
  \begin{subfigure}{0.32\linewidth}
    \includegraphics[width=1.0\linewidth]{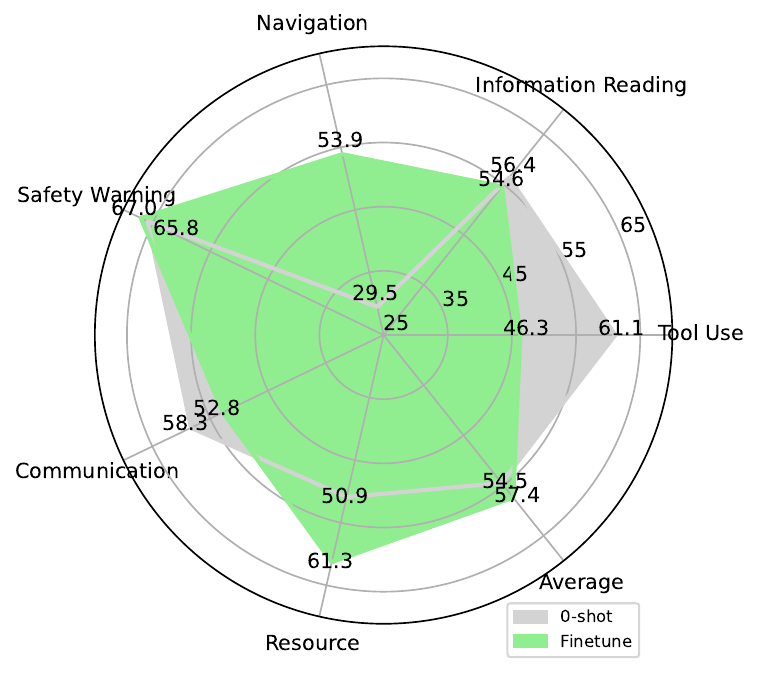}
    \caption{LLaVA-OV.}
    \label{fig:it_gemini}
  \end{subfigure}
   \begin{subfigure}{0.32\linewidth}
    \includegraphics[width=1.0\linewidth]{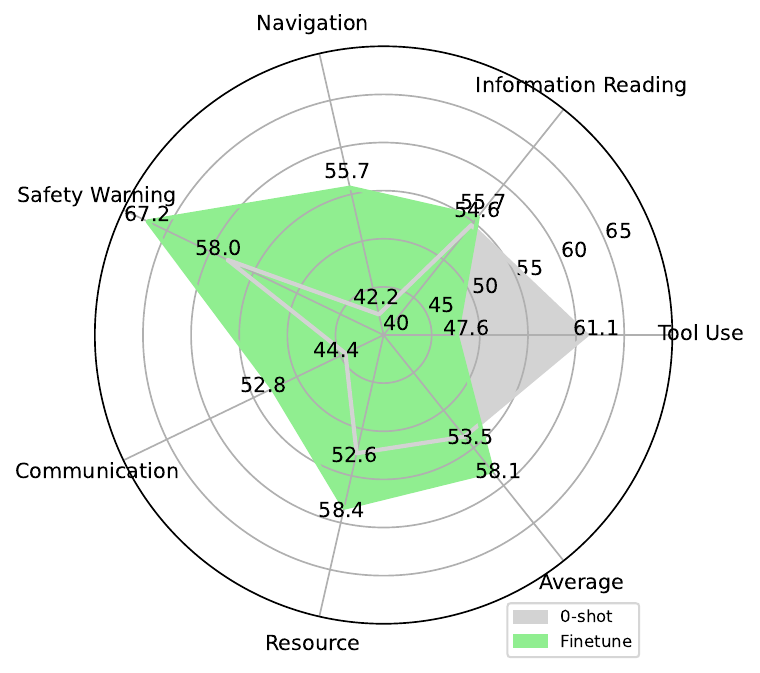}
    \caption{InternVL-2.5-8B.}
    \label{fig:it_gpt4}
  \end{subfigure}
  \caption{Simple finetuning with \dataset~training data can notably boost QA performances, but it is ill-suited for answering ``tool-use'' questions.}
  \label{fig:aba-inst}
  \vspace{-3mm}
\end{figure}

\subsection{Other Investigations}
Our investigations aim to answer three questions: 
1) Is a single frame at the question timestamp sufficient to answer the question?
3) Does explicitly prompting models to assist blind users benefit performance?
2) Can instruction tuning with our training data improve performance?

Table \ref{tab:sigframe} shows that replacing multi-frame inputs with a single frame at the question timestamp degenerates model performances, though results remain competitive (which is reasonable as people often ask about their current visual environment.). Figure \ref{fig:aba-vf} further suggests that video-level modeling is key for answering questions of ``Safety Warning'', ``Tool Use'', and ``Communication'', while a single frame seems to be sufficient for information reading. Table \ref{tab:blindpromt} shows that substituting the blind-specific prompt with a normal VQA prompt jeopardizes model performance, highlighting the unique challenges of blind-oriented QA, especially in navigation and tool using as shown in Figure \ref{fig:aba-prompt}. Table \ref{tab:itune} and Figure \ref{fig:aba-inst} demonstrate that using our training data for instruction tuning (LoRA \cite{hu2022lora} finetune) can remarkably improve performances, yet the gap compared with human remains significant (Implementation details are presented in Appendix Sec. \ref{sec:inv}). Additionally, Figure \ref{fig:aba-inst} shows that existing finetuning will hurt the performances on ``tool use'' questions, likely because of overfitting on the limited training data for this category. 
\section{Conclusion}
We present the construction of a VideoQA dataset \dataset~to reflect on the challenges towards egocentric visual assistance for the blind. \dataset~is the first of its kind in that the videos are taken by the visually-impaired individuals from a first-person perspective, with questions posed and verified by blind users to ensure close alignment with their true assistant needs. We further comprehensively benchmark the challenge with multiple prominent multimodal LLMs and unveil their significant limitations in various aspects. We conduct thorough analysis and share many key insights for future advancements in this direction. By formulating and bringing the challenge to the vision-language community, our primary goal is to push the MLLM research towards live egocentric visual assistance for the overwhelming number of visually impaired people around the world.

Importantly, we have ensured that the dataset collection and user studies adhere to IRB standards. Informed consent was obtained from all the content creators, who agreed to their content being used for non-commercial research purposes. Access to the dataset will be 
granted with the condition that video sources will be cited for distribution. 




\newpage
\bibliography{ref}
\bibliographystyle{ieeetr}




\newpage
\appendix
\section*{Appendix}
\label{sec:app}

\section{\dataset~Dataset Creation}
\label{sec:appdataset}

\subsection{QA Generation with GPT-4o}
\label{sec:aigen}
To enrich the QA diversity and reduce the annotation burden, we prompt GPT-4o to act as blind users to generate part of questions and invite both blind and sighted people to check and edit them. For generation, we first decode the video into frames at 3 fps. We then adopt an adaptive sampling strategy over the decoded frames to obtain the final frames to be fed to GPT-4o for QA generation. We sample at a ratio of every 18 frames for long videos (\eg, 180s) and 6 frames for short ones (\eg, 30s). To simulate real-time QA generation, each frame is accompanied by its timestamp and a refined prompt (shown in Table \ref{tab:gen_qa}).
The generation process is divided into batches by question categories. This means that each batch contains only one type of question, which can help in organizing and enriching the dataset, ensuring a wide coverage of different topics and improving overall quality. 
The specific prompts for each category are attached in Table \ref{tab:gen_qa}.

\subsection{Blind User Study}
\label{sec:blindstudy}
\begin{wrapfigure}[14]{r}{0.49\textwidth}
    \centering 
    \includegraphics[width=1.0\linewidth]{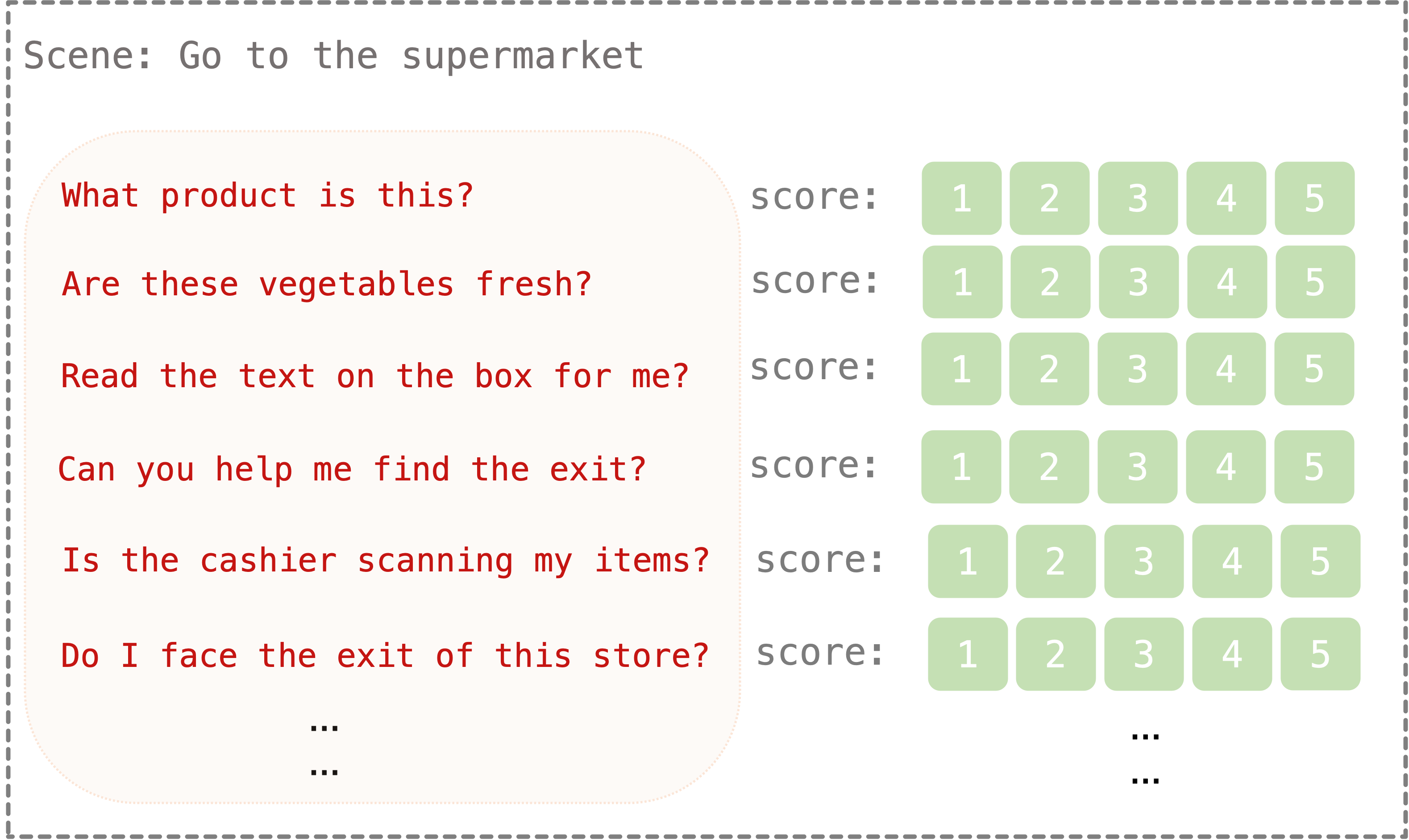} 
    \caption{The scoring example} 
    \label{fig:example} 
\end{wrapfigure}

\begin{figure*}[t!]
    \centering
    \includegraphics[width=1.0\linewidth]{ 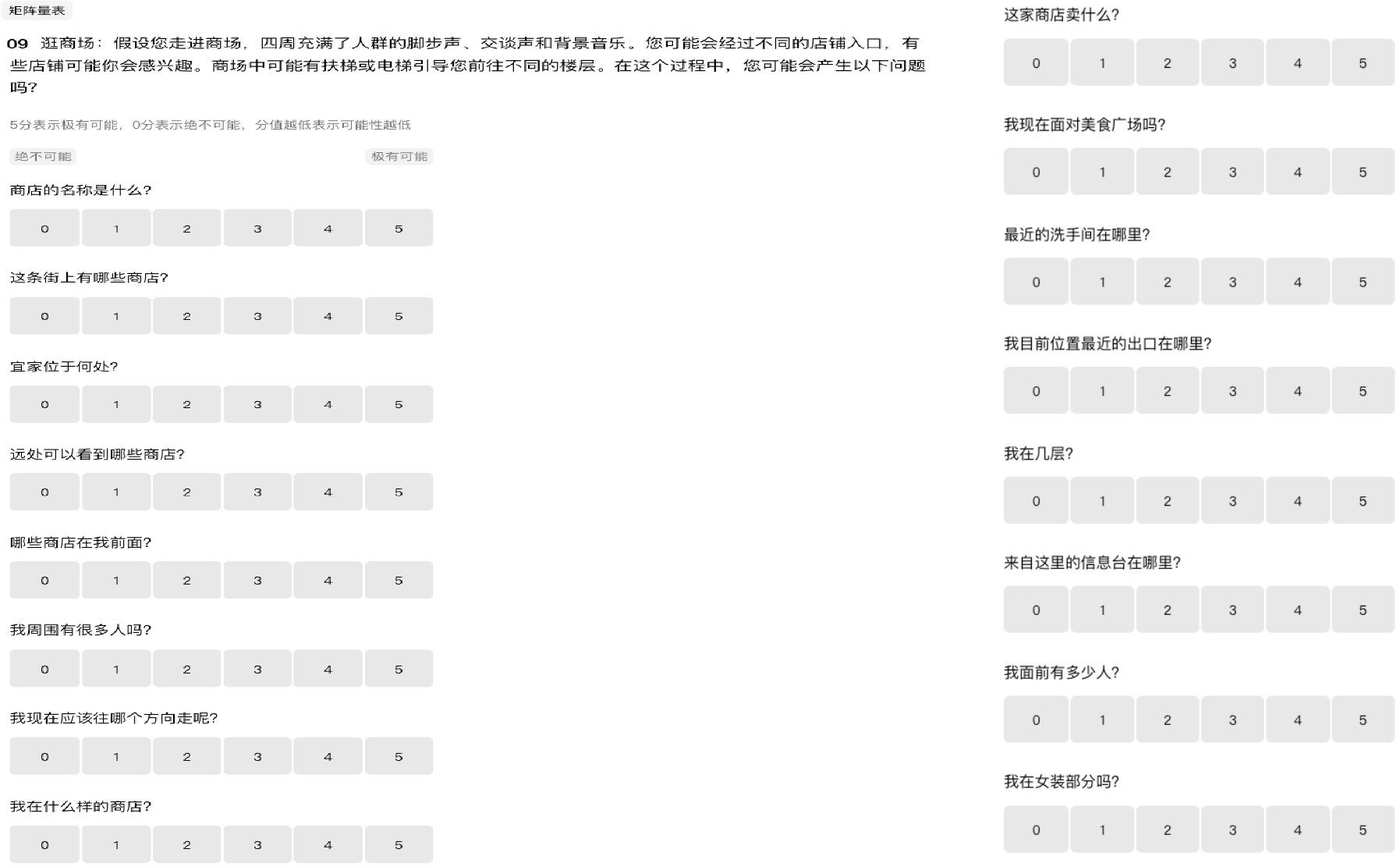}
    \caption{Exampe of questionnaire for blind study. Questions are translated into Chinese.}
    \label{fig:questionnaire}
    \vspace{-0.3cm}
\end{figure*}

We categorize manually corrected QA pairs (translated into Chinese) based on video scenes and invite 111 blind individuals to evaluate the \textbf{blind-relevance} of the questions across different scenarios. The evaluation scale ranges from 0 to 5, with higher scores indicating preferable questions and lower scores reflecting unwanted questions (as illustrated in Figure \ref{fig:example}, a real example is shown in Figure \ref{fig:questionnaire}). Additionally, blind individuals are encouraged to provide further questions and insights to better meet the needs of individuals with visual impairments.

\begin{figure}[h!]
  \centering
  \begin{subfigure}{0.49\linewidth}
    \includegraphics[width=1.0\linewidth]{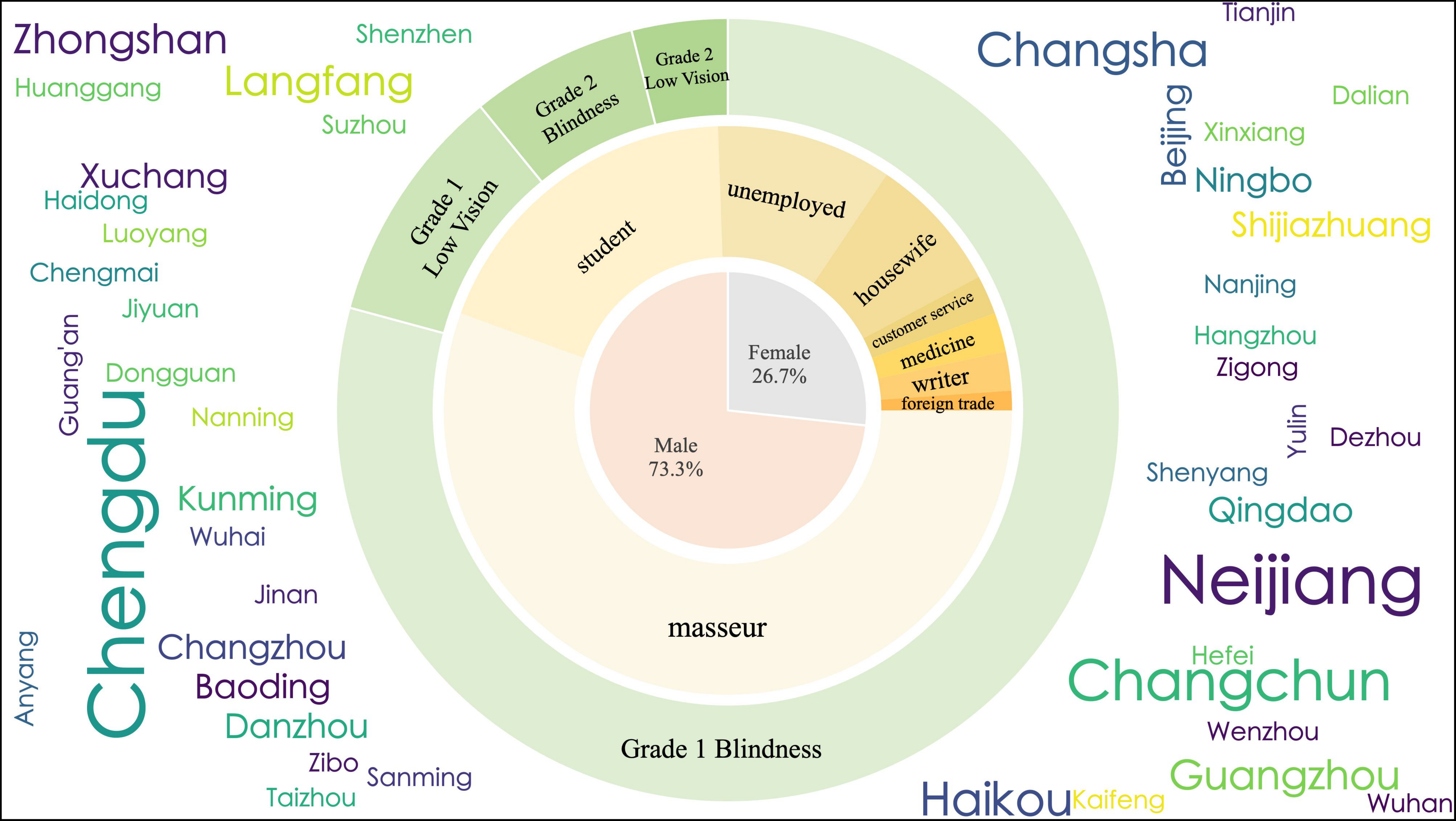}
    \caption{Location distribution.}
    \label{fig:local}
  \end{subfigure}
  \begin{subfigure}{0.49\linewidth}
    \includegraphics[width=1.0\linewidth, height=0.57\textwidth]{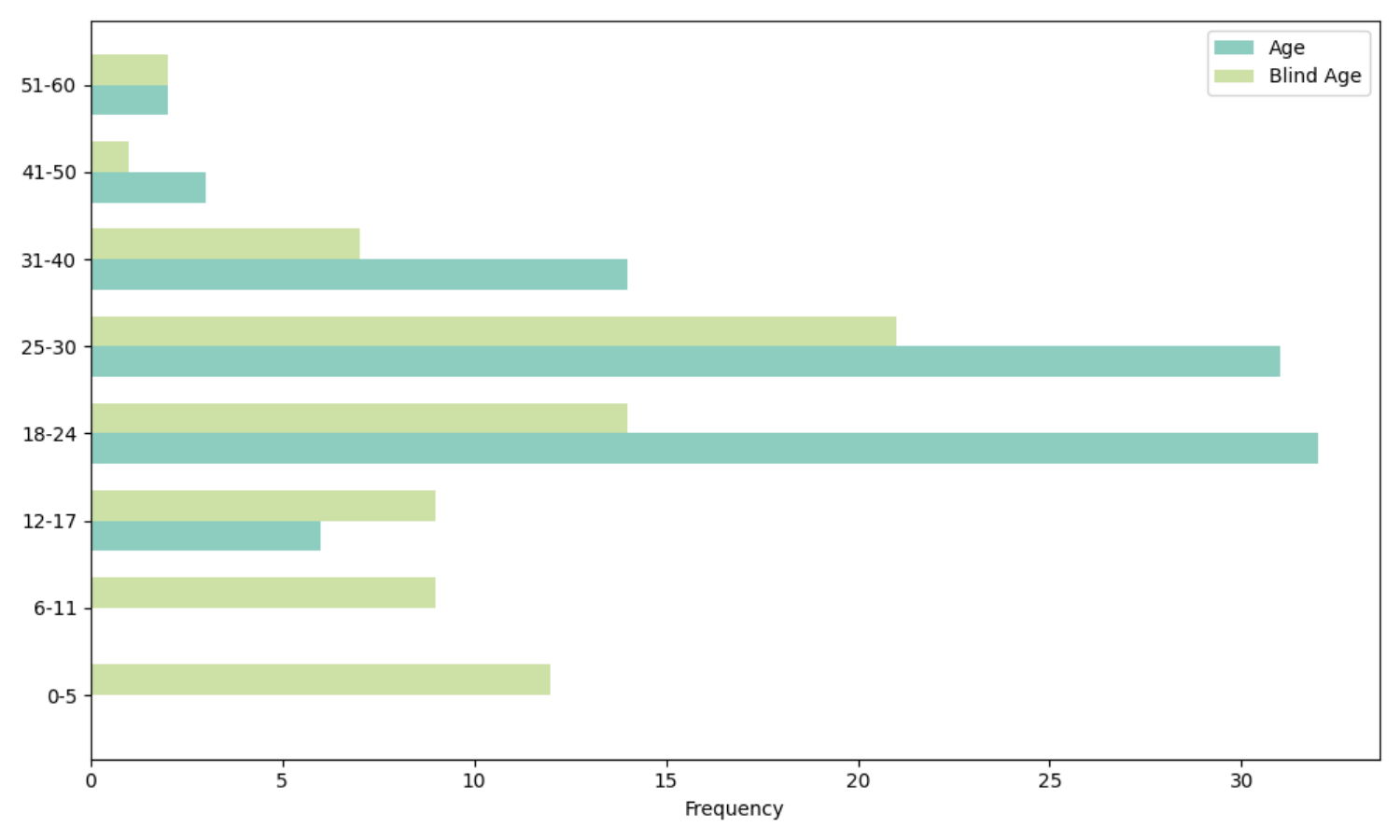}
    \caption{Age distribution.}
    \label{fig:age}
  \end{subfigure}
  \caption{Distribution of blind participants}
  \label{fig:blinduser}
  \vspace{-0.4cm}
\end{figure}

The cohort study (Figure \ref{fig:local}) exhibits substantial demographic heterogeneity, encompassing participants ranging from teenagers to older adults (Figure \ref{fig:age}). Male participants account for a dominant proportion of 73.3\%, with young adults aged 18-30 constituting the largest age cohort. Notably, for most respondents, the duration of visual impairment corresponded with their biological age, suggesting congenital origin - a finding consistent with the observation that over 80\% presented Grade 1 Blindness (total visual acuity $\leq$0.02 or visual field $<5$\textdegree). The remaining participants acquire visual impairment postnatally, though all reported living with the condition for long time.

The occupational distribution reveals a pronounced concentration in specialized massage services (55.6\%), reflecting a predominant vocational pattern among the visually impaired population of China. Students form the secondary occupational group (18.9\%), followed by unemployed individuals (10.0\%) and other professions (15.5\%).  Geographically, the sample represents 46 urban centers across China's hierarchical city-tier system, ranging from first-tier megacities (Beijing, Guangzhou, Shenzhen) to emerging technology hubs (Hangzhou, Chengdu, Changsha) and smaller urban cities (Neijiang, Zhongshan, Langfang).

This demographic and geographic diversity ensures comprehensive coverage of life scenes across varying urban contexts, from high-density metropolitan networks to community-level systems.

\begin{figure}[!t] 
    \centering 
    \includegraphics[width=1.0\textwidth]{ 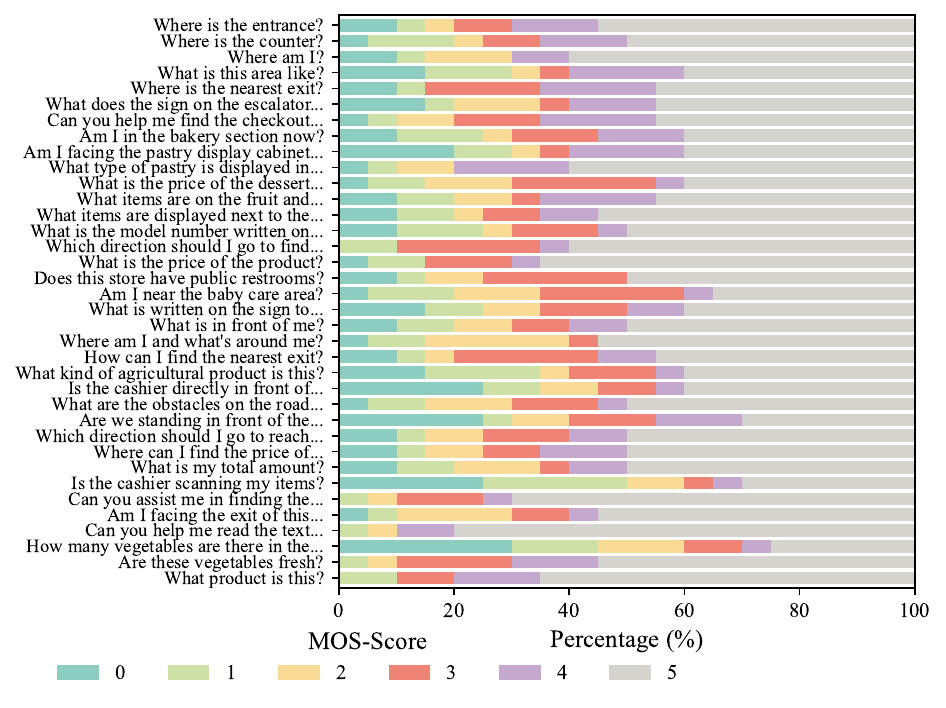} 
    \caption{MOS score distribution of blind people in shopping scenarios. The stacked parts of each color represent the distribution proportions of different scores (same for below).}
    \label{fig:supermarket} 
    \vspace{-0.3cm}
\end{figure}

We found that when blind people visit supermarkets, stores, and shopping malls (as shown in Figure \ref{fig:supermarket}); their primary concerns are navigation and positioning issues related to location. Specifically, they want to know where they are and how to get to their desired destination(``How to get to XXX from here?") . Their second main concern is obtaining product information, such as price (``Where can I find the price of this product?"), style, shelf life, etc. Additionally, some blind individuals are also interested in understanding the range of products sold at specific stores and the layout of the mall, including questions like ``What does this store sell?" and ``Which products are displayed on the counter?"
At the same time, we found that blind individuals tend to be less concerned about the presence and location of people around them. Questions related to the number of people nearby (``How many people are in front of me?") and their specific whereabouts (``Is the cashier directly in front of me?") received a significantly higher proportion of low scores compared to high scores. Additionally, they showed minimal interest in information about road obstacles. The proportion of low scores for questions such as "What are the obstacles on the road?" and "How many floors are there?" was considerably higher than the proportion of high scores.

\begin{figure}[!t] 
    \centering
    \includegraphics[width=1.0\textwidth]{ 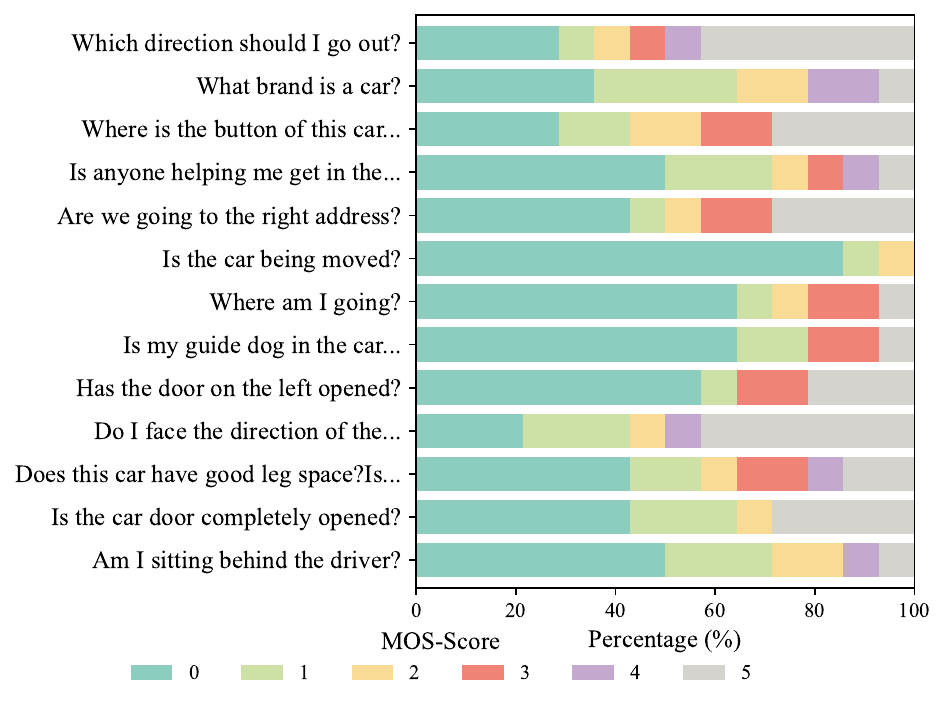} 
    \caption{MOS score distribution of blind people using transportation.} 
    \label{fig:transport}
    \vspace{-0.3cm}
\end{figure}

\begin{figure}[!t] 
    \centering 
    \includegraphics[width=1.0\linewidth]{ 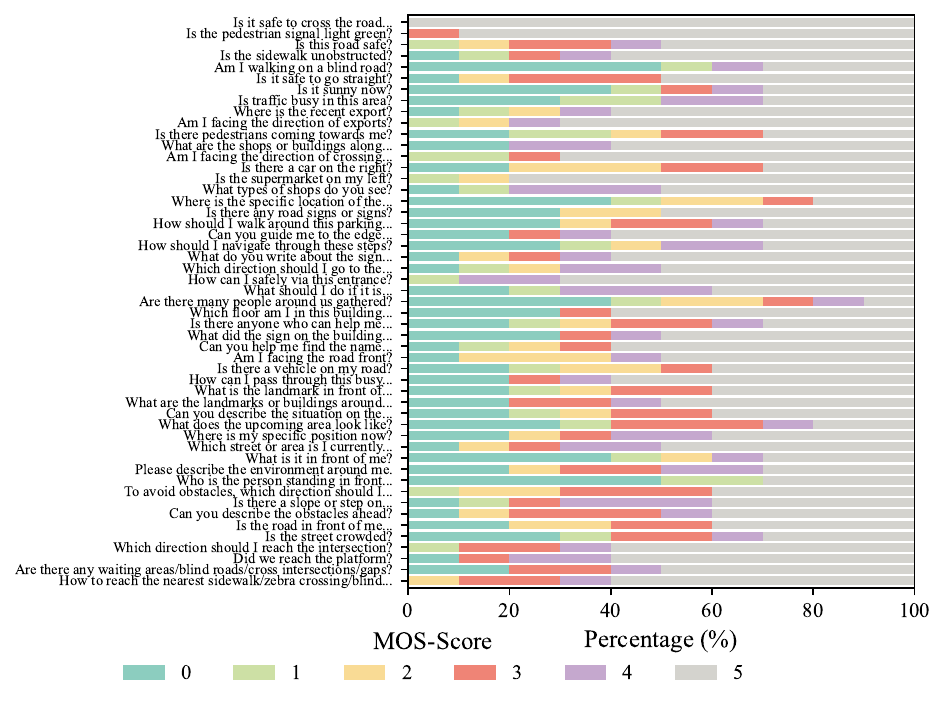} 
    \caption{MOS score distribution of blind people navigating outdoor environments. } 
    \label{fig:goout} 
    \vspace{-0.3cm}
\end{figure}

\begin{figure}[!t] 
    \centering 
    \includegraphics[width=0.9\linewidth]{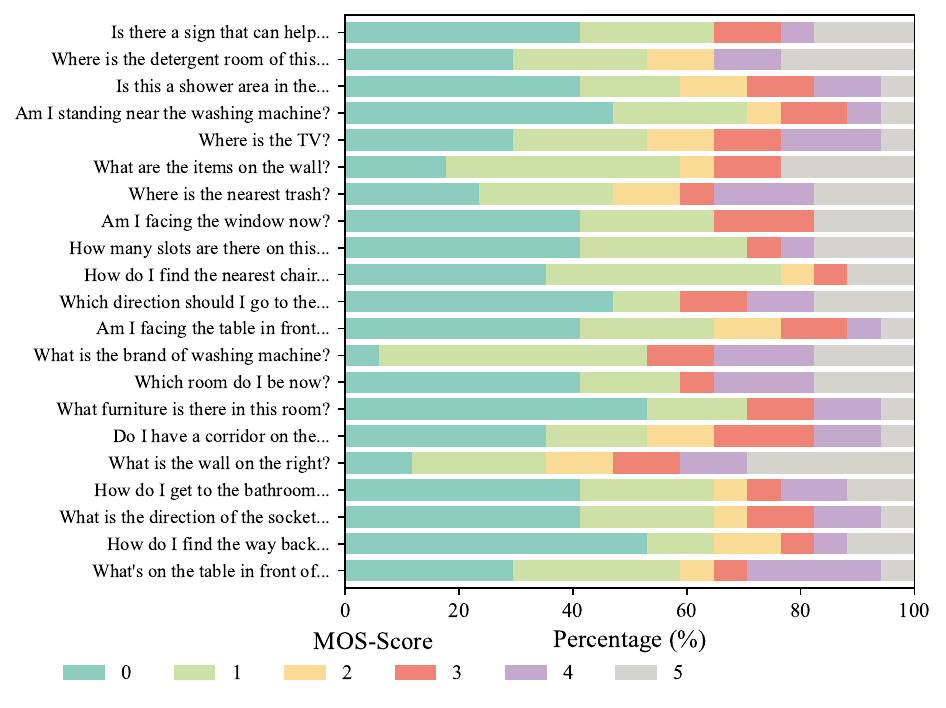} 
    \caption{MOS score distribution of blind people at home. } 
    \label{fig:home} 
    \vspace{-0.3cm}
\end{figure}

\begin{figure}[!t] 
    \centering 
    \includegraphics[width=1.0\linewidth]{ 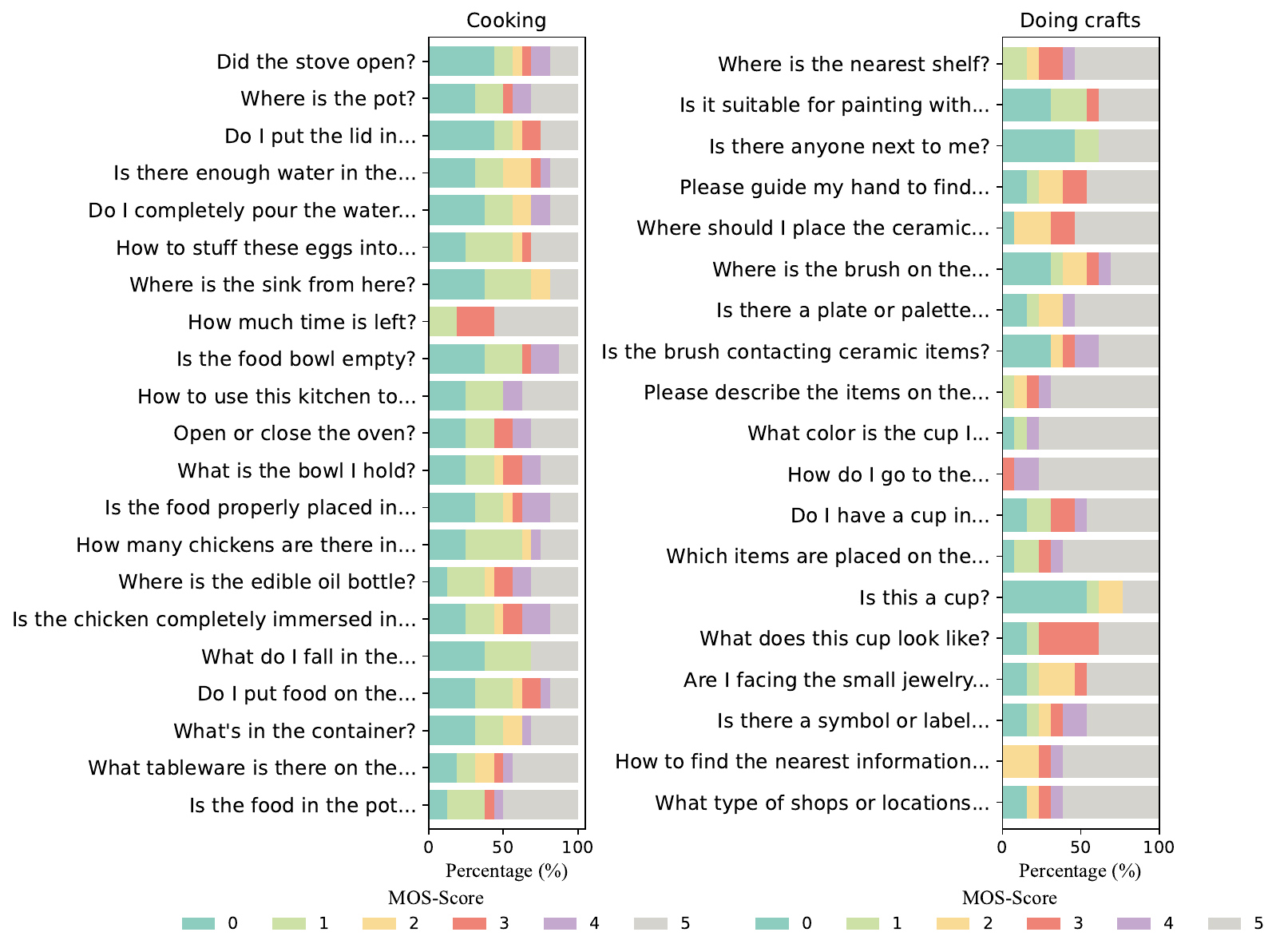} 
    \caption{MOS score distribution of blind people in cooking and doing craft scenarios.} 
    \label{fig:cook_craft}
    \vspace{-0.3cm}
\end{figure}

When blind people use transportation such as cars, buses, and other vehicles (as shown in Figure \ref{fig:transport}), over 60\% of the respondents assigned full scores to questions like "Is this my taxi?" and "Which bus is coming?" while waiting for a taxi. This indicates that their primary concern is the current location of the vehicle. Meanwhile, respondents uniformly gave low scores to the question "Is the car being moved?" suggesting that blind individuals are capable of perceiving vehicle movement. Furthermore, approximately 80\% of respondents gave low scores to questions such as "What does the bus stop advertisement say?" and "What brand is the car?" indicating that blind people often show little interest in irrelevant information surrounding their waiting position.

In outdoor navigation scenarios, such as finding destinations, crossing roads, or walking, blind individuals prioritize safety above all else (as shown in Figure \ref{fig:goout}). For instance, the question ``Is it safe to cross the road?" receives the highest score of 5 from all respondents. Similarly, over 80\% of respondents rate questions related to safety during movement, such as ``How can I safely get through this entrance?" and ``Is it safe to go straight?", with scores of 4 or higher.
Blind individuals also place significant emphasis on directional and positional information. For questions about accessing public facilities, such as ``How to reach the nearest sidewalk, zebra crossing, or blind road?", over 70\% of respondents gave scores of 4 or higher. Direction-related questions like ``Which direction should I take to reach the intersection?" and ``Is the supermarket on my left?" received scores above 3 from more than 60\% of respondents.

However, respondents show less concern for questions about the characteristics of the surrounding environment, such as ``Please describe the environment around me" and ``Is traffic busy in this area?". Similarly, questions about weather conditions (``Is it sunny now?") or information about nearby pedestrians (``Who is the person standing in front?", ``Are there many people around us gathered?", ``Is there a pedestrian coming towards me?") received scores below 3 from more than 50\% of respondents, indicating a general lack of interest.
Interestingly, questions about obstacles did not receive as high a score as might be intuitively expected. For example, fewer than 50\% of respondents positively rated questions like ``Can you describe the obstacles ahead?" and ``To avoid obstacles, which direction should I go?", while a large proportion gave neutral scores (3).  This may be due to differences in the proficiency of using white canes among blind individuals, which influences their reliance on such information. Similarly, questions such as ``Am I walking on a blind road'' also exhibited a polarized response, with respondents either assigning very low scores (0 or 1) or very high scores (4 or 5).
This polarization suggests that reliance on certain types of information may vary significantly depending on individual preferences and skills.

We also observe characteristics of blind individuals’ activities in familiar environments that contradict common assumptions. As shown in Figure \ref{fig:home}, when blind individuals are at home, we found that the proportion of questions receiving low scores (0, 1, or 2) exceeded 60\% across almost all queries. Respondents explained that although they cannot see, their brains form a mental map of the environment based on prior familiarity. This mental representation allows them to navigate and plan their movements seamlessly within the environment. Unless there are significant changes in the environment, they typically do not encounter any issues.

To validate the rationality of question design in hands-on activity scenarios for visually impaired individuals, we conducted a study where participants rated questions in scenarios such as cooking and crafting (As shown in Figure \ref{fig:cook_craft}). Taking these two scenarios as examples, we found that, despite the differences in context, respondents consistently gave high scores (4 and 5) to core questions that directly influence task completion. For instance, “How much time is left?” (Cooking) and ``Please describe the items on the shelves'' (Doing crafts).
At the same time, due to the differences in activities, visually impaired individuals tended to give a higher proportion of low scores (0 or 1) to cooking-related questions compared to crafting. This can be attributed to the fact that cooking typically occurs in familiar environments where blind individuals rely on their spatial memory and experience. In contrast, in the crafting scenario, respondents showed a neutral or low level of interest (0–3) in questions about basic item information such as appearance or shape, exemplified by questions like ``Is it a cup?'' and ``What does this cup look like?''.
\begin{figure}[t!]
    \centering 
    \includegraphics[width=1.0\linewidth]{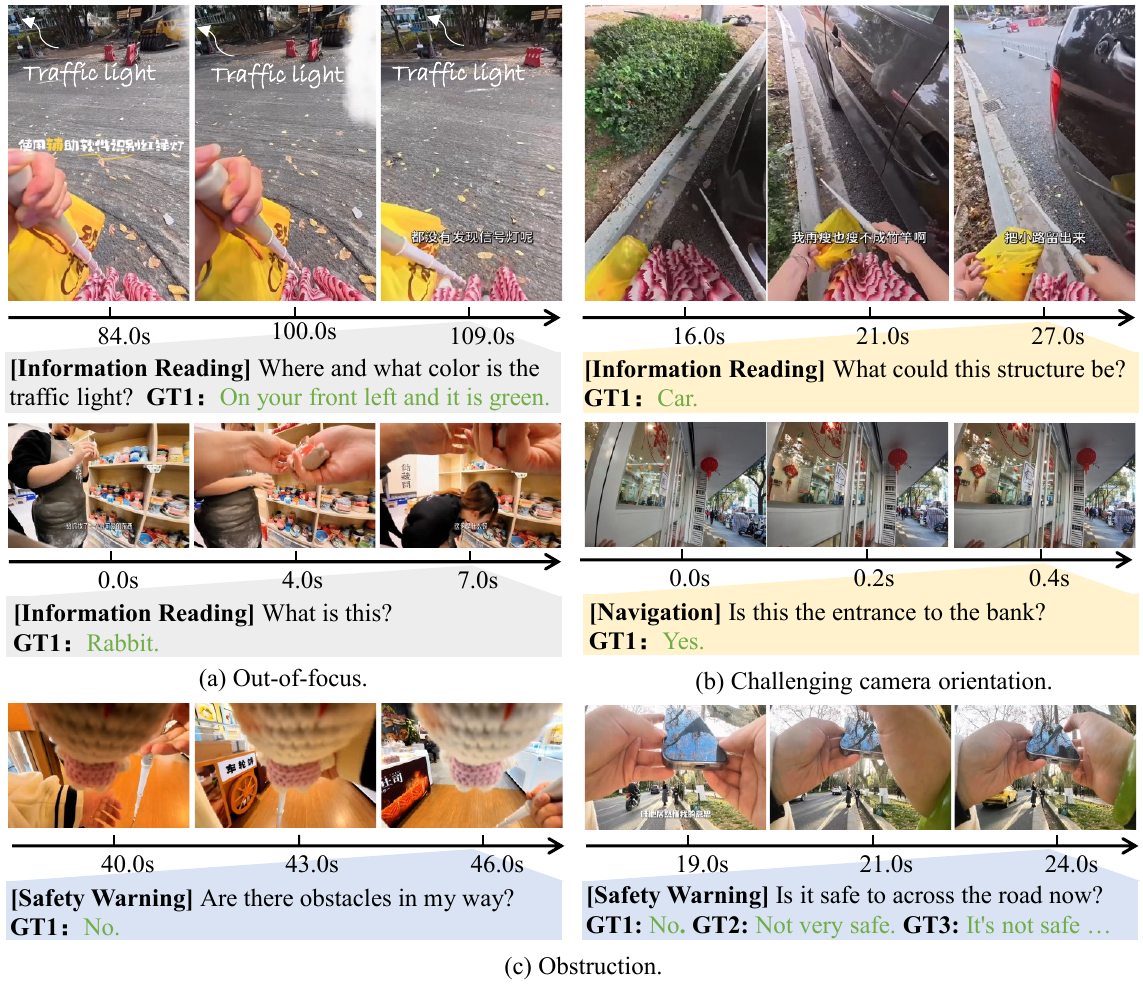} 
    \caption{Visualization of EgoBlind examples.}
    \label{fig:video}
    \vspace{-0.3cm}
\end{figure}

\subsection{Analysis of Videos}
\label{sec:compvideo}
We analyze more details about the videos in EgoBlind. First, unlike the famous Ego4D \cite{grauman2022ego4d} videos which are framed in a top-down view of sighted people via head-mounted camera devices (\eg, smart glasses), the videos in \dataset~are captured by GoPros mounted in front of the blind users' chests
or mobile phones held in their  hands. 
Specific challenges resulting from EgoBlind videos are: 1) The objects of interest are often off-center and out-of-focus due to blindness, as shown in Figure \ref{fig:video}(a), 2) the objects are shown in challenging viewpoint (blind users tend to walk alongside a street curb or a building for safety.), and 3) The scene could be largely occluded by their canes, mobile phones, guide dogs or other people.

\section{Experiment}
\label{sec:appexp}

\subsection{Evaluation Metric}
\label{sec:appeval}
\begin{wrapfigure}[16]{r}{0.36\textwidth} 
    \centering 
    \vspace{0cm}
    \includegraphics[width=1.0\linewidth]{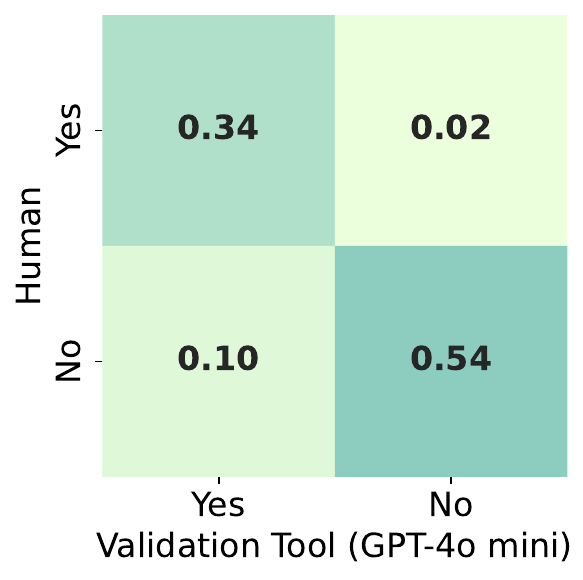} 
    \caption{Consistency between human and machine scoring.}
    \label{fig:consistency}
\end{wrapfigure}
To enhance the reliability of the AI evaluation tool
(\ie, GPT-4o mini), we sample 30\% of the test data to obtain inference results of Gemini 2.0. We invite human volunteers to score the models’ inference results on this subset. Based on the results of human scoring, we refine the prompts to guide the machine’s evaluation results closer to the human scoring outcomes.
Ultimately, we calculate the Cohen Kappa coefficient \cite{kappa2012} (for inter-rater reliability) between human and GPT scoring results, which is found to be 0.73, indicating a high degree of consistency between the two. Through heatmap analysis (shown in Figure \ref{fig:consistency}), we observe that samples where human and machine ratings are consistent accounted for 88\% of the total test data. For samples where ratings are inconsistent, we find that most of they are due to an information bias -- machine cannot see the videos for answer scoring. Our final evaluation prompts are attached in Table \ref{tab:evalprompt}.

\begin{figure}[t!]
    \centering 
    \includegraphics[width=1.0\linewidth]{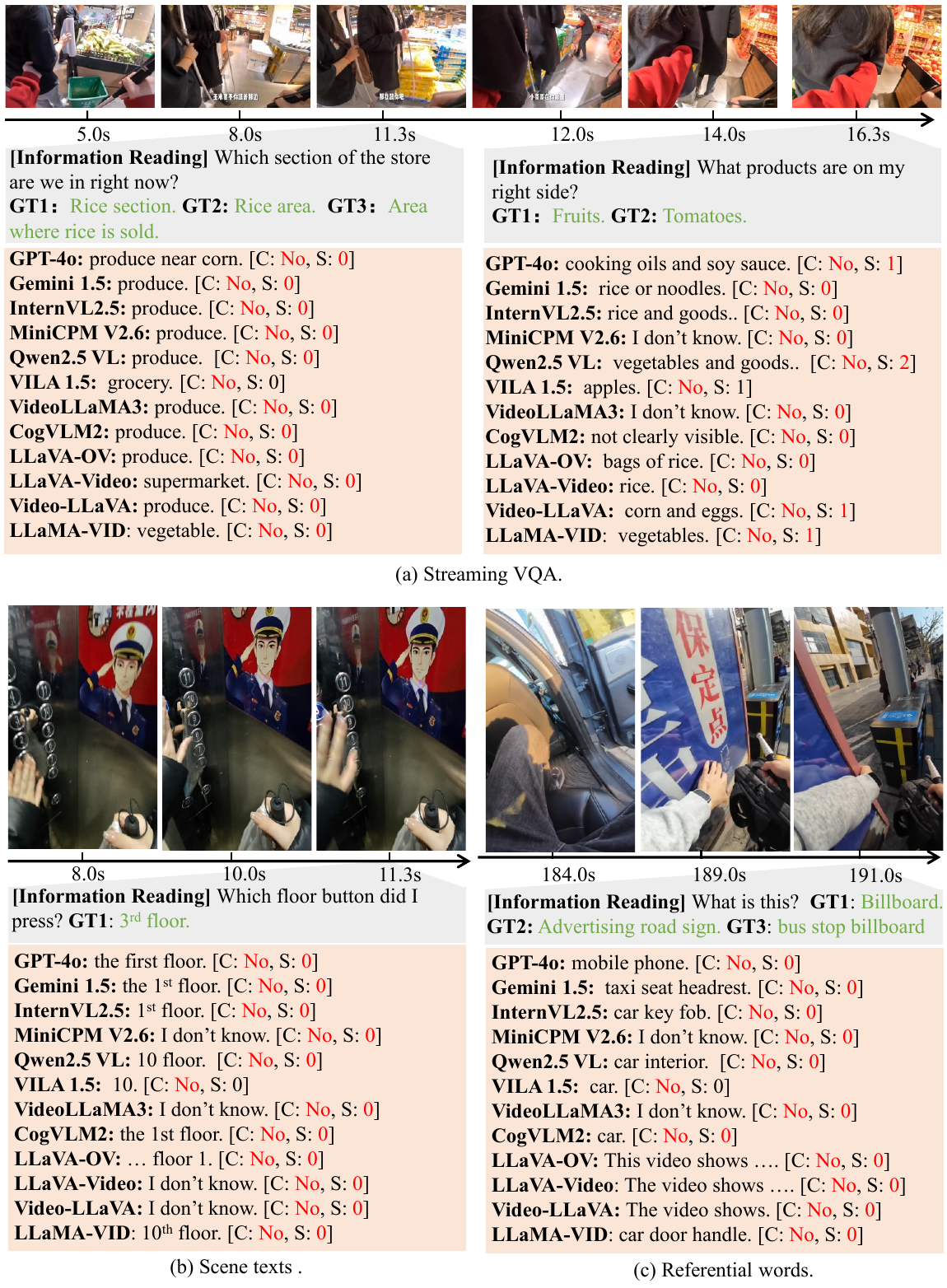} 
    \caption{Visualization of failure cases. C: Correctness, S: Score. We only show the key answer words for brevity (same for below).}
    \label{fig:otherchallenge}
    \vspace{-0.3cm}
\end{figure}
\subsection{Benchmark Analysis}
\label{sec:casestudy}

\textbf{Other Common Challenges.}
Figure \ref{fig:otherchallenge} shows other common failure cases of the current Video-LLMs towards egocentric visual assistance for the blind. The example in Figure \ref{fig:otherchallenge}(a) demonstrates that the models are hard to perform effective live QA with streaming visual inputs. They often rely on the wrong frames for responses, even though we have explicitly prompted that the question is posed at the last frame (refer to our prompts at Table \ref{tab:blind-prompt}). The example in Figure \ref{fig:otherchallenge}(b) indicates that the models are poor at scene text recognition, as all model fail to answer the 3rd floor which was pressed by the user. The example also shows that the models are weak in performing temporal context reasoning. Because human can easily know the answer as the user moves her hand upward from button "1" to the 3rd button, even though the button "3" is partly blocked by the hand. Finally, the example in Figure \ref{fig:otherchallenge}(c) suggests that the models are struggling in linking the referential words (\eg,``this") with the visual contents. This also necessitates spatio-temporal context understanding, \eg, following the hand motion and the object it interacted with.

\subsection{Other Investigations}
\label{sec:inv}

\begin{table}[t!]
\centering
\small
\caption{Overall QA accuracy (\%) of user-agonistic data split \vs user-aware data split.}
\begin{threeparttable}
\begin{tabular}{l|cc|cc}
    \toprule
     \multirow{2}{*}{Models} &  \multicolumn{2}{c|}{User-Agnostic} & \multicolumn{2}{c}{User-Aware} \\ 
     & Zero-shot & Finetune & Zero-shot & Finetune \\ 
    \midrule
    Qwen2.5-VL & 45.5 & 50.2 \textcolor{green}{$\uparrow$ 4.7} & 47.7 & 52.9 \textcolor{green}{$\uparrow$ 5.2} \\
    LLaVA-OV &  54.5 & 57.4 \textcolor{green}{$\uparrow$ 2.9} & 56.4 & 62.0 \textcolor{green}{$\uparrow$ 5.6} \\
    InternVL2.5-8B & 53.5 & 58.1 \textcolor{green}{$\uparrow$ 4.6} & 56.2 & 62.3 \textcolor{green}{$\uparrow$ 6.1} \\
    \bottomrule
\end{tabular}
\end{threeparttable}
\label{tab:usersplit}
\vspace{-4mm}
\end{table}

\begin{table}[t!]
\centering
\small
\caption{Results of category-specific prompting for each question group (half of test data).}  
\label{tab:category_prompt}  
\resizebox{\textwidth}{!}{  
\begin{tabular}{@{}lccccccc|l@{}}  
\toprule  
\textbf{Method} & \multicolumn{1}{c}{\textbf{category prompt}} & \textbf{Tool} & \textbf{Info.} & \textbf{Nav.} & \textbf{Safe} & \textbf{Com.} & \textbf{Res.} & \textbf{Overall} \\
\midrule  
\multirow{2}{*}{LLaVA-OV} &  \xmark & 58.4  & 54.1 &  37.2  &  63.8 & 59.4  & 54.0 &  54.2  \\
 & \checkmark & 50.6  & 53.2 & 30.2 & 64.1 & 56.2 & 52.9 & 52.2 \textcolor{red}{$\downarrow$ 2.0} \\
\midrule
\multirow{2}{*}{InternVL2.5-26B} & \xmark & 74.0  & 56.0 &  47.7  &  51.9 & 46.9 & 56.3 &  54.6  \\
 & \checkmark & 66.2  & 55.5 & 52.8 & 43.0 & 53.1 & 63.2 & 53.1 \textcolor{red}{$\downarrow$ 1.5} \\
\midrule
\multirow{2}{*}{GPT-4o} & \xmark &  61.0  & 59.6 &  54.3  &  60.3 & 46.9 &  69.0 &  59.4  \\
 & \checkmark & 79.2  & 63.3 & 58.8 & 58.6 & 50.0 & 64.4 & 62.2 \textcolor{green}{$\uparrow$ 2.8} \\
\bottomrule  
\end{tabular}
}
\end{table}

\begin{table}[t!]
\centering
\small
\caption{Investigation of Chinese-specific elements (half of test data). Subtitles and scene texts (SText) play little role in EgoBlind, while all model performance decline significantly when translating the questions into Chinese (CHN). }  
\label{tab:ch_factor}  
\resizebox{\textwidth}{!}{  
\begin{tabular}{@{}lccccccccc|l@{}}  
\toprule  
\textbf{Method} & \multicolumn{1}{c}{\textbf{Subt.}} & \textbf{SText} & \textbf{CHN} & \textbf{Tool} & \textbf{Info.} & \textbf{Nav.} & \textbf{Safe} & \textbf{Com.} & \textbf{Res.} & \textbf{Overall} \\
\midrule  
\multirow{4}{*}{Qwen2.5-VL} &  \checkmark & & & 45.4  &  49.0  & 32.2 & 46.5 &  40.6 & 35.6 &  44.5 \\
 &  \xmark & & & 44.2 & 46.3 & 27.6 & 51.0 & 40.6 & 31.0 & 43.2 \textcolor{red}{$\downarrow$ 1.3} \\
 &  \checkmark &  \checkmark &  & 42.9 & 48.5 & 33.7 & 45.5 & 46.9 & 43.7 & 44.8 \textcolor{green}{$\uparrow$ 0.3} \\
  &  \checkmark &  & \checkmark & 44.2 & 46.4	& 31.2	& 40.1	& 31.2 & 39.1 & 41.5 \textcolor{red}{$\downarrow$ 3.0} \\
\midrule
\multirow{4}{*}{LLaVA-OV} & \checkmark & & & 58.4  & 54.1 &  37.2  &  63.8 & 59.4  & 54.0 &  54.2  \\
 & \xmark & &  & 52.0  &54.4 & 34.2 & 64.4 & 59.4 & 55.2 & 53.7 \textcolor{red}{$\downarrow$ 0.5} \\
 & \checkmark & \checkmark & & 52.0 & 56.4 & 35.2 & 62.5 & 53.1 & 54.0 & 54.1 \textcolor{red}{$\downarrow$ 0.1} \\
  &  \checkmark &  & \checkmark & 52.0 & 53.0 & 36.7 & 60.3 & 40.6 & 46.0 & 51.4 \textcolor{red}{$\downarrow$ 2.8} \\
\midrule
\multirow{4}{*}{InternVL2.5-26B} &\checkmark & & & 74.0  & 56.0 &  47.7  &  51.9 & 46.9 & 56.3 &  54.6  \\
 & \xmark & & &67.5  & 52.5 & 51.3 & 53.8 & 50.0 & 57.5 & 53.8 \textcolor{red}{$\downarrow$ 0.8} \\
 & \checkmark & \checkmark & & 62.3 & 57.4 & 48.7 & 49.7 & 53.1 & 55.2 & 54.2 \textcolor{red}{$\downarrow$ 0.4} \\
 &  \checkmark &  & \checkmark & 59.7& 56.0 & 48.7 & 50.3 & 50.0	& 49.4 & 53.1 \textcolor{red}{$\downarrow$ 1.5} \\
\midrule
\multirow{4}{*}{GPT-4o} & \checkmark & & & 61.0  & 59.6 &  54.3  &  60.3 & 46.9 &  69.0 &  59.4  \\
 & \xmark & & & 68.8  & 56.9 & 53.8 & 55.8 & 53.1 & 70.1 & 57.6 \textcolor{red}{$\downarrow$ 1.8} \\
 & \checkmark & \checkmark & & 63.6 & 59.0	& 50.8 & 53.2 & 56.2 & 62.1	& 56.7 \textcolor{red}{$\downarrow$ 2.7} \\
 &  \checkmark &  & \checkmark & 64.9 & 55.1 & 51.8 & 56.4 & 56.2 & 60.9	& 55.9 \textcolor{red}{$\downarrow$ 3.5} \\
\bottomrule  
\end{tabular}
}
\end{table}

\textbf{Implementation Details for Finetuning.}
The fine-tuning procedures are conducted on two NVIDIA A800 GPUs. Most hyper-parameters are kept at their default configurations as specified in the fine-tuning scripts. For InternVL2.5-8B, optimal performance is achieved by increasing the training epochs to 2 while finetuning only the MLP layer. Regarding LLaVA-OV, we employ LoRA with reduced rank dimensionality (r=16) to optimize computational efficiency. Due to constraints of compute resource, we truncate the maximum sampled frames to 16, and cape the model's maximum sequence length to 8192, and freeze the vision encoder. Other refinements involve extending the number of epochs to 2, increasing gradient accumulation steps to 4, and adjusting weight decay to 0.05. Finally, for Qwen2.5-VL, we find that the best results are achieved by simply training the model 1 epoch while preserving other parameters unchanged.

\textbf{Category-aware Prompting.}
We investigate whether designing category-specific prompt for each question group can enhance QA performance. To reduce API costs for closed-source models, the experiments are conducted on a randomly selected half of the test set. Results in Table \ref{tab:category_prompt} show that category-specific prompting (see Table \ref{tab:ques-prompt}) effectively improves GPT-4o's performance, but not for other open-source models. We speculate that the open-source models are not strong enough to understand longer and complex prompts.

\textbf{User-aware Dataset Split.}
Additionally, we re-split our EgoBlind dataset into train and test sets by ensuring videos token by the same user not appearing in both, \ie, half users in train and half users in test. We obtain both the zero-shot and finetuned results of three representative open-source models on the new test set in Table \ref{tab:usersplit}. The results show that finetuning can remarkably improve the model performance under different data splits.
To our surprise, all models perform better with the user-aware split, regardless of zero-shot or finetuning. We speculate that some user videos and questions are not that challenging to answer.

\textbf{Chinese Contents.}
While EgoBlind aims for ``in-the-wild'' visual assistance for the blind, the videos are majorly sourced from Chinese video sharing platforms. Hence, we study if the model performances are sensitive to Chinese-specific elements in the videos. Specifically, we first remove all subtitles from videos by using off-the-shelf subtitle remover \footnote{https://github.com/YaoFANGUK/video-subtitle-remover}. Results in Table \ref{tab:ch_factor} show that all model performances drop slightly by 0.5\% to 1.8\%. A manual check of the new failure cases reveals that the vast majority (95.5\% to 98.4\% for different models) are irrelevant to subtitles and cloud be due to visual input variation resulted from subtitle removal (\eg, visual blur), not because of no subtitles. Alternatively, we extract scene texts from videos by using advanced OCR tool\footnote{https://github.com/PaddlePaddle/PaddleOCR}. The scene texts are used as additional prompting contents for MLLMs to make decisions. The results in Table \ref{tab:ch_factor} show that the additional scene texts do not help model predictions but slightly hurt the performance. The results and analyses suggest that Chinese subtitles and scene texts matter little in answering EgoBlind questions. Finally, we translate all questions into Chinese and find that all model performance decline significantly, reflecting that all models are weaker in understanding Chinese compared to English.

\section{Limitations}
While we take significant challenge to collect the videos and QAs, the data are limited to the mainland of China so far. Also, the training set is not that big compared with other video QA datasets from sighted people, since there is not much egocentric video data from the blind online. However, we have shown the effectiveness and importance of the training data for better performances. Also, we are continuously collecting related data and aim to extend the dataset by cooperating with the blind association, possibly enclosing blind VQA data from all over the world. In addition, the MLLM techniques are evolving rapidly and we may have missed some most recent models for testing. We will setup an evaluation server and maintain a leader-board to trace the technique advancements in egocentric visual assistance for the blind.

\section{Societal Impacts}
\textbf{Positive Impacts:} \dataset~is the first VideoQA dataset that benchmarks vision-language research towards egocentric visual assistance for the blind.  The research and outcomes can support developing assistive technology for blind and visually impaired individuals. \dataset~can advance intelligent systems capable of understanding and describing real-world environments from a first-person perspective, thereby enhancing autonomy, safety, and quality of life for blind users. 
By enabling more natural and interactive communication between the blind users and the assistive system, VideoQA technologies can help bridge accessibility gaps in daily activities such as navigation, information reading, tool use, social interaction and other resource acquisition, thus holding immerse value towards enlightening the blind individual's life.

\textbf{Negative Impacts:} 
Despite its significance. Privacy concerns may arise as egocentric recordings inherently capture sensitive and personal information about both the blind user and bystanders. Additionally, biases in training data or system limitations may lead to inaccurate or misleading responses, potentially compromising user trust or safety. Addressing these challenges is essential to ensure the ethical and inclusive deployment of MLLMs in practical visual assistance for the blind.

\clearpage

\begin{table}[t!]
\centering
\small
\caption{Normal prompts for models to perform question answering on \dataset.}
\begin{tabular}{|l|p{10cm}|}
\midrule
\textbf{Model} & \textbf{General Prompts} \\ \midrule
InternVL2.5-26B & I will provide you with a video each time and one question; Your task is to answer the question which was posed by the people in the last frame of the video. The answer needs to be based on the content of the picture and the objective characteristics. If the question cannot be answered, you can say I don't know. Do not include Chinese characters in your response. The question is: {row['question']} \\ \midrule

GPT-4o & I will provide you with several pictures each time and one question; Your task is to answer the question. The answer needs to be based on the content of the picture and the objective characteristics. If the question cannot be answered, you can say I don't know. Please generate the response with keys 'question', 'question\_id', 'prediction', and 'type'. The question is: \{question\}, Question ID: \{question\_id\}, Type: \{question\_type\}. The content of the key 'question' is the question you received. The content of the key 'prediction' is the answer to the corresponding question you generated. The key 'question\_type' represents which category the question type you received. Your response should look like this example: \{'question':'Is there a wet floor caution sign near the door?', 'question\_id': 'v\_87\_1\_2', 'prediction':'yes.', 'type': 'information reading'\} \\ \midrule
Gemini 1.5 Flash & I will provide you with a video each time and one question; Your task is to answer the question which was posed by the people in the last frame of the video. The answer needs to be based on the content of the picture and the objective characteristics. If the question cannot be answered, you can say I don't know. Do not include Chinese characters in your response. Please generate the response with keys 'question', 'question\_id', 'prediction', and 'type'. The question is: \{question\},Question ID: \{question\_id\},Type: \{question\_type\}. The content of the key 'question' is the question you received. The content of the key 'prediction' is the answer to the corresponding question you generated. The key 'question\_type' represents which category the question type you received. Your response should look like this example: \{\{"question":"Is there a wet floor caution sign near the door?", "question\_id": "v\_87\_1\_2", "prediction":"yes.", "type": "information reading"\}\} \\ \midrule

\end{tabular}
\label{tab:normal-prompt}
\end{table}

\begin{table}[t!]
\centering
\small
\caption{Blind-aware prompts for models to perform question answering on \dataset.}
\begin{tabular}{|l|p{10cm}|}
\midrule
\textbf{Model} & \textbf{General Prompts} \\ \midrule
Open Source MLLMs & I will provide you with a video each time and one question; These questions are all questions raised by the blind person in the video from his own first-person perspective in the current scene. Your task is to answer the blind person's question which was posed in the last frame of the video. The answer needs to be based on the content of the picture and the objective characteristics that the blind person cannot see. If the question cannot be answered, you can say I don't know. Do not include Chinese characters in your response. The question is: \{question\} \\ \midrule

GPT-4o & I want you to act as a visual assistant for the blind. I will provide you with several pictures each time and one question; These questions are all raised by the blind person in the video from his own first-person perspective in the current scene. Your task is to answer the blind person's question. The answer needs to be based on the content of the picture and the objective characteristics that the blind person cannot see. If the question cannot be answered, you can say I don't know. Please generate the response with keys 'question', 'question\_id', 'prediction', and 'type'. The question is: \{question\},Question ID: \{question\_id\},Type: \{question\_type\}. The content of the key 'question' is the question you received. The content of the key 'prediction' is the answer to the corresponding question you generated. The key 'question\_type' represents which category the question type you received. Your response should look like this example: \{'question':'Is there a wet floor caution sign near the door?', 'question\_id': 'v\_87\_1\_2', 'prediction':'yes.', 'type': 'information reading'\} \\ \midrule

Gemini & I want you to act as a visual assistant for the blind. I will provide you with a video each time and one question; These questions are all questions raised by the blind person in the video from his own first-person perspective in the current scene. Your task is to answer the blind person's question which was posed in the last frame of the video. The answer needs to be based on the content of the picture and the objective characteristics that the blind person cannot see. If the question cannot be answered, you can say I don't know. Do not include Chinese characters in your response. Please generate the response with keys 'question', 'question\_id', 'prediction', and 'type'. The question is: \{question\},Question ID: \{question\_id\},Type: \{question\_type\}. The content of the key 'question' is the question you received. The content of the key 'prediction' is the answer to the corresponding question you generated. The key 'question\_type' represents which category the question type you received. Your response should look like this example: \{"question":"Is there a wet floor caution sign near the door?", "question\_id": "v\_87\_1\_2", "prediction":"yes.", "type": "information reading"\} \\ \midrule

\end{tabular}
\label{tab:blind-prompt}
\end{table}

\begin{table}[t!]
\centering
\small
\caption{Category-specific prompts. They are properly inserted into the QA prompts in Table \ref{tab:blind-prompt}.}
\begin{tabular}{|l|p{10cm}|}
\midrule
\textbf{Question Category} & \textbf{Prompts} \\ \midrule
Tool Use & The questions focus on tool use, helping the blind understand how to operate tools around them. \\ \midrule

Information Reading & The questions focus on information reading, helping the blind understand visible facts such as what their surroundings look like. \\ \midrule

Navigation & The questions focus on navigation, guiding the blind to move safely or find directions in their environment. \\ \midrule

Safety Warnings & The questions focus on safety warnings, reminding the blind of possible dangers of their surroundings. \\ \midrule

Social Communication & The questions focus on social communication, recognizing or describing people interacting with the blind. \\ \midrule

Resource & The questions focus on other resources, identifying people or things nearby that may offer help. \\ \midrule
\end{tabular}
\label{tab:ques-prompt}
\end{table}

\begin{table}[ht]
\centering
\caption{Prompts for GPT-4o mini to evaluate MLLMs on \dataset.} 
\begin{tabular}{@{}p{12cm}@{}}
\toprule
\textbf{Evaluation Prompts} \\
\midrule
You are an intelligent chatbot designed for evaluating the correctness of generative outputs for question-answer pairs. Your task is to compare the predicted answer with the correct answer and determine if they match meaningfully. Here's how you can accomplish the task: 

\textbf{\#\#INSTRUCTIONS:} 

- \textbf{Focus on meaningful matches:} Assess whether the predicted answer and the correct answer have a meaningful match, not just literal word-for-word matches. 

- \textbf{Criteria for Correctness:} The predicted answer is considered correct if it reasonably matches any of the four standard answers, recognizing that synonyms or varied expressions that convey the same meaning are acceptable. 

- \textbf{Allow for Paraphrasing:} Understand that different wording that conveys the same fundamental idea is valid. Evaluate if the essence of the predicted answer captures the core information of the correct answer. 

- \textbf{Flexibility in Evaluation:} Use judgment to decide if variations in the predicted answer still correctly address the question, even if they do not directly replicate the correct answer's phrasing. 

For example, when the correct answer is 'Left front', Predicted Answer: 'About ten meters to your left front', these two answers match. 

Please evaluate the following video-based question-answer pair: 

\texttt{Question: \{question\}} \\
\texttt{Correct Answer0: \{answer0\}} \\
\texttt{Correct Answer1: \{answer1\}} \\
\texttt{Correct Answer2: \{answer2\}} \\
\texttt{Correct Answer3: \{answer3\}} \\
\texttt{Predicted Answer: \{pred\}} 

Provide your evaluation only as a yes/no and score where the score is an integer value between 0 and 5, with 5 indicating the highest meaningful match. Please generate the response in the form of a Python dictionary string with keys \texttt{'pred'} and \texttt{'score'}, where the value of \texttt{'pred'} is a string of 'yes' or 'no' and the value of \texttt{'score'} is in INTEGER, not STRING. 

DO NOT PROVIDE ANY OTHER OUTPUT TEXT OR EXPLANATION. Only provide the Python dictionary string. For example, your response should look like this: \texttt{\{'pred': 'yes', 'score': 4.8\}}. \\
\bottomrule
\label{tab:evalprompt}
\end{tabular}
\end{table}

\begin{table}[ht]
\centering
\caption{Example prompts for GPT-4o to generate QAs.} 
\begin{tabular}{@{}p{12cm}@{}}
\toprule
\textbf{Social Communication} \\
\midrule
I want you to act as a blind person. I will provide you with several pictures each time and the pictures have two characteristics: 1. pictures have temporal correlation between them and the order of the pictures represents the order of time. 2.The pictures are all taken from your first point of view, which is equivalent to the picture recorded by the camera hanging on your chest.

Your task is to imagine what questions you would ask to meet your needs if you were in the situation of the picture provided. You need to focus on asking questions about ‘communication and interaction’. Below I will give a detailed explanation of ‘communication and interaction’.

Communication and Interaction: Describe the interaction between blind people and surrounding people/companions, and observe other people's status information for multi-person collaborative activities. For example: Did my companion help me carry my luggage onto the conveyor belt? /Has my guide dog entered the elevator safely? /Who is talking to me?

Here’s how you can accomplish the task: 1. use your imagination and only ask about ‘communication and interaction’ questions; 2.Please ask questions about each picture; 3. The questions you ask need to be real-time and do not contain any off-site information; 4.The questions you ask need to be answered through picture content; 5.If you cannot answer based on the picture, you can answer ‘I don't know’; 6. Ask as many open-ended questions as possible, not just 'yes-no' questions; 7. Some of the questions generated need to reflect the characteristics of the video, that is, you need to watch several consecutive frames to get the answer. For example:'What is my companion doing? 8. ‘communication and interaction’ questions should be based on the real life scenarios of blind people, and the questions must be practical (that is, questions that blind people would really ask in the current situation). 9. Please don't ask bad examples like ‘What is my companion saying to me ?’/‘Is there an announcement about the next station or any other important information?’

Please generate the response with 'timestamp', 'question', 'answer' and  'questoin\_type'.
The content of the key 'timestamp' the time value corresponding to the current picture, for example: the first image corresponds to the first element in timestamp list, and so on. The {timestamp} value can be found here: 'f' Timestamp list: {time}.\\
\bottomrule
\label{tab:gen_qa}
\end{tabular}
\end{table}

\end{document}